\documentclass[conference]{IEEEtran}
\IEEEoverridecommandlockouts
\usepackage{cite}
\usepackage{amsmath,amssymb,amsfonts}
\usepackage{algorithmic}
\usepackage{graphicx}
\usepackage{textcomp}
\usepackage{xcolor}
\usepackage{algorithm}
\usepackage{algorithmic}
\usepackage{epsfig}
\usepackage{amsthm,amsmath,amssymb}
\usepackage{makecell}
\usepackage{mathrsfs}
\usepackage{multirow}
\usepackage{multicol}
\usepackage{booktabs}
\usepackage{xcolor}  
\usepackage{url} 
\newcommand\blfootnote[1]{%
  \begingroup
  \renewcommand\thefootnote{}\footnote{#1}%
  \addtocounter{footnote}{-1}%
  \endgroup
}
\def\BibTeX{{\rm B\kern-.05em{\sc i\kern-.025em b}\kern-.08em
    T\kern-.1667em\lower.7ex\hbox{E}\kern-.125emX}}
\begin{document}

\title{Detecting AI-Generated Video via Frame Consistency}

\author{\textbf{
   Long Ma$^{1}$,  Zhiyuan Yan$^{2}$,  Qinglang Guo$^{1}$,
 Yong Liao$^{1\dag}$,  Haiyang Yu$^{3\dag}$,  Pengyuan Zhou$^{4}$ }\\

 $^{1}$ \textmd{School of Cyber Science and Technology, University of Science and Technology of China}\\
 $^{2}$ \textmd{School of Electronic and Computer Engineering, Peking University} \\
 
 $^{3}$ \textmd{School of Software Technology, Zhejiang University}
 $^{4}$\textmd{Aarhus University}\\
 \{longm@mail, yliao@\}ustc.edu.cn
}
\maketitle

\blfootnote{$^\dagger$ Corresponding Author. This work was supported in part by Key Science \& Technology Project of Anhui Province 202423l10050033. }
\begin{abstract}
The increasing realism of AI-generated videos has raised potential security concerns, making it difficult for humans to distinguish them from the naked eye. Despite these concerns, limited research has been dedicated to detecting such videos effectively.
To this end, we propose an \textit{open-source AI-generated video detection dataset}. 
Our dataset spans diverse objects, scenes, behaviors, and actions by organizing input prompts into independent dimensions. 
It also includes various generation models with different generative models, featuring popular commercial models such as OpenAI's Sora, Google's Veo, and Kwai's Kling.
Furthermore,  
we propose a simple yet effective \textit{\underline{De}tection model} based on \underline{Co}ncistency of \underline{F}rame (DeCoF), which learns robust temporal artifacts across different generation methods.
Extensive experiments demonstrate the generality and efficacy of the proposed DeCoF in detecting AI-generated videos, including those from nowadays' mainstream commercial generators.
\end{abstract}


\section{Introduction}
\label{Introduction}

Existing generative models such as Diffusion models~\cite{nichol2021improved,rombach2022high,parmar2023zero} have excelled in generating high-quality visual contents, driving advancements in video generation~\cite{blattmann2023align,zhang2023show} that produce highly realistic videos capable of deceiving the human eye.
Unfortunately, these advancements have raised significant security concerns, such as privacy violations and the erosion of trust on social media~\cite{sharma2023survey,barrett2023identifying}, underscoring the urgent need for effective and reliable detection of AI-generated videos.

A straightforward solution is to use and fine-tune the existing spatiotemporal neural networks (STNNs) for detection, which involves I3D~\cite{carreira2017quo}, SlowFast~\cite{feichtenhofer2019slowfast}, and Mvit~\cite{fan2021multiscale}, conventionally designed for video tasks like video understanding. 
These pre-trained STNNs intuitively seem capable of capturing both spatial and temporal forgery artifacts for AI-generated video detection. 
However, our initial investigation reveals that naively training an STNN can lead to a shortcut, where the model relies on easily recognizable spatial artifacts while neglecting the more subtle, potentially generalizable temporal artifacts. 
This shortcut results in generalization issues when detecting videos from unseen generators (see results in Tab.~\ref{stcnn}). 

\begin{figure}[t] 
\centering
\includegraphics[width=0.95\columnwidth]{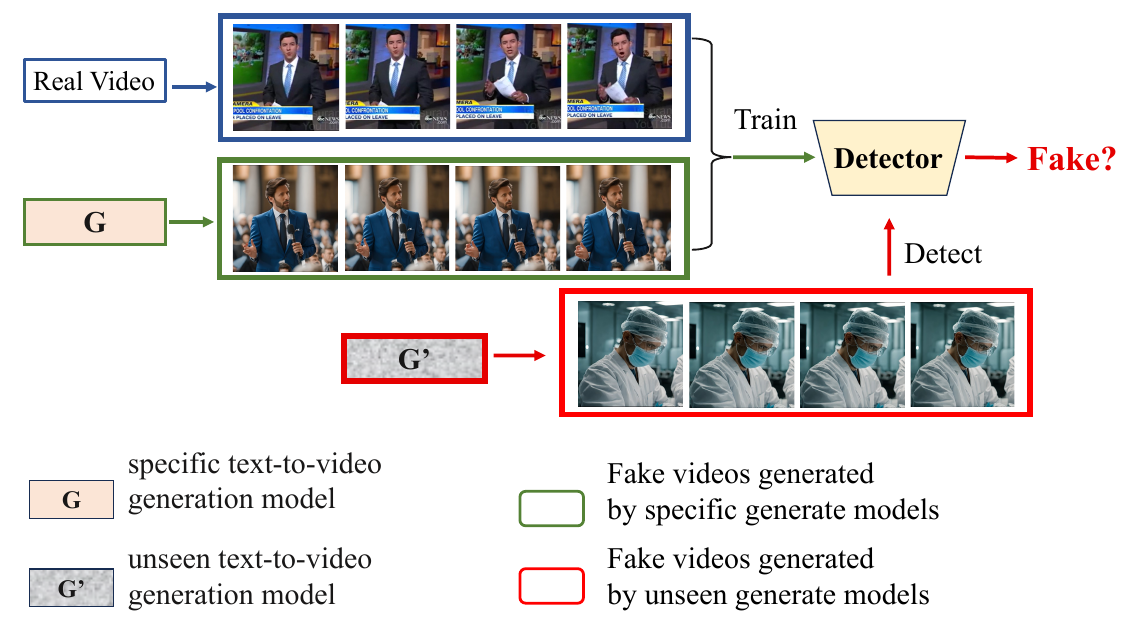}
\caption{Can the detector trained on a specific video generation model detect videos generated by unseen generation models?}
\label{fig:th1}
\end{figure}

To address these problems, in this paper, we have proposed (1) a comprehensive \textbf{AI-generated video detection dataset} with various detection methods and aligned content settings; and (2) a \textbf{robust temporal detector} based on temporal forgery artifacts.
\textbf{First}, we propose an AI-generated video detection dataset, i.e., the $Generated$ $Video$ $Forensics$ (GVF) dataset, which is constructed to evaluate the performances of AI-generated video detectors. 
GVF covers a wide range of scene content and motion changes through careful selection of prompts, simulating diverse real-world environments.
We carefully select a series of prompts and retrieve corresponding real videos. 
Fake videos are then created using the same prompts, forming real-fake pairs that encourage the detector to focus on forgery artifacts rather than content differences.
Furthermore, we have used \textbf{nine T2V generators} with different generation methodologies (e.g., pixel-space or latent-space models), comprehensively evaluating the performance of the detection methods. 
\textbf{Second}, we propose a robust detection method relying on temporal artifacts. Specifically, we decouple temporal and spatial artifacts by mapping video frames to a feature space where the inter-feature distance is inversely correlated with image similarity, enabling the detection of anomalies caused by inter-frame inconsistencies. Our method significantly reduces computational complexity and memory requirements, as it focuses solely on learning anomalies between features.
Extensive experiments demonstrate that our method exhibits extraordinary capabilities in AI-generated video detection and excellent generalizability for unseen generation models, even on the latest generation models Sora~\cite{Sora}, Veo~\cite{Veo} and Kling~\cite{Kling}.

Our contributions can be summarized as follows:
\begin{itemize}
    \item We present the first comprehensive and scalable dataset for benchmarking AI-generated video detectors. 
    \item We develop a robust AI-generated video detector named DeCof that aims to learn the general temporal artifacts.
    \item We validate the effectiveness of our approach through extensive experiments, including tests on the latest commercial proprietary models.
\end{itemize}

\section{Related work}
\label{Related work}
\subsection{Generation Models}
Recently, with the success of diffusion models in generating images \cite{ho2020denoising,song2020denoising,nichol2021improved,esser2021taming}, a number of research explores text-to-video Diffusion Models (VDMs). Some works~\cite{ramesh2022hierarchical,blattmann2023align,khachatryan2023text2video,wang2023modelscope} ground their models on latent-based VDMs, e.g., Text2Video-zero~\cite{khachatryan2023text2video} generates videos by using motion dynamics to enhance the latent code of text-to-image(T2I) model~\cite{rombach2022high} without the need for training on any video data, and ModelScopeT2V~\cite{wang2023modelscope} extends T2I model with temporal convolution and attention blocks trained with image-text and video-text datasets. ZeroScope\cite{zero} is a watermark free version of ModelScopeT2V, and has been optimized for 16:9 composition. On the other hand, PYoCo~\cite{ge2023preserve} and Make-A-Video~\cite{singer2022make} anchor their models on pixel-based VDMs. Show1~\cite{zhang2023show} integrates Pixel-based VDMs to produce videos of lower resolution with better text-video alignment, then upscale them with latent-based VDMs to create high-resolution videos with low computation cost.

\subsection{Generated Content Detection}
Many studies have focused on generated image detection, especially for unseen generation models. A simple classifier trained by Wang et al.~\cite{wang2020cnn} on ProGAN-generated images can be generalized to 20 unseen other GAN-generated images. Frank et al.~\cite{frank2020leveraging} conclude that the artifacts of the generated images are caused by the upsampling method. Wang et al.~\cite{wang2023dire} identify generated images through the difference of images reconstructed by a pre-trained diffusion model. 
Ojha et al.~\cite{ojha2023towards} propose to develop a universal generated image detector based on pre-trained vision-to-language models.
Meanwhile, a large amount of work focuses on Deepfake  detection~\cite{shiohara2022detecting}.
So far, there are some studies on detecting Deepfake videos \cite{yan2023ucf,yan2024df40,yan2024transcending,yan2023deepfakebench}, 
but research on detecting AI-generated videos remains unexplored. This paper targets at bringing exploratory and leverageable contributions to this new direction.

\begin{figure}[t]

\begin{minipage}[b]{0.24\linewidth}
  \centerline{\epsfig{figure=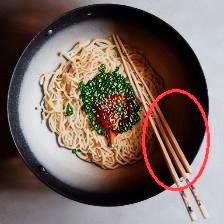,width=2cm}}
  \centerline{(a) Geometry}\medskip
  
\end{minipage}
\hfill
\begin{minipage}[b]{.24\linewidth}
  \centerline{\epsfig{figure=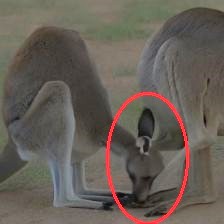,width=2cm}}
  \centerline{(b) Layout}\medskip
\end{minipage}
\hfill
\begin{minipage}[b]{0.24\linewidth}
  \centerline{\epsfig{figure=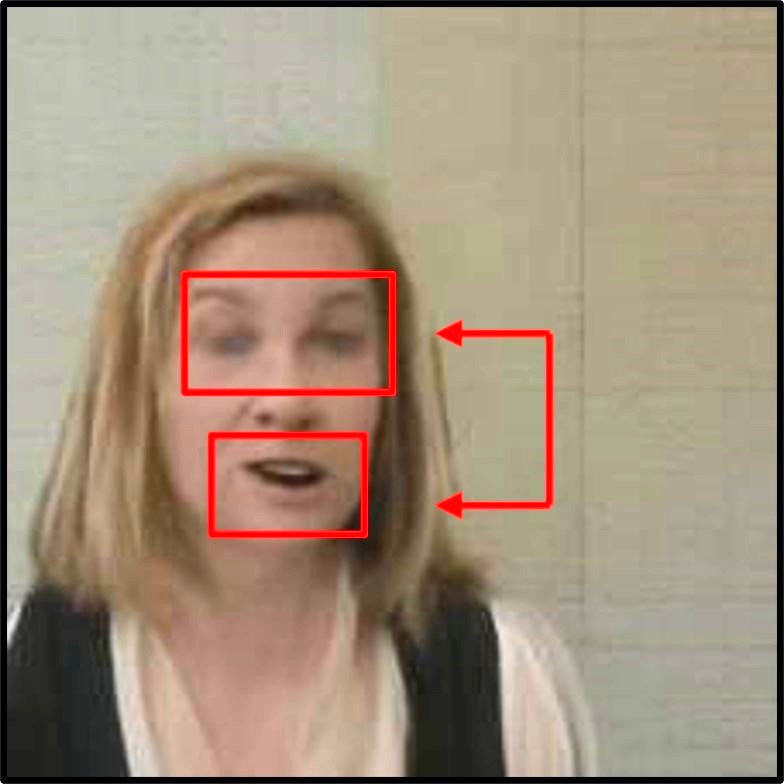,width=2cm}}
  \centerline{(c) Frequency}\medskip

\end{minipage}
\hfill
\begin{minipage}[b]{0.24\linewidth}
  \centerline{\epsfig{figure=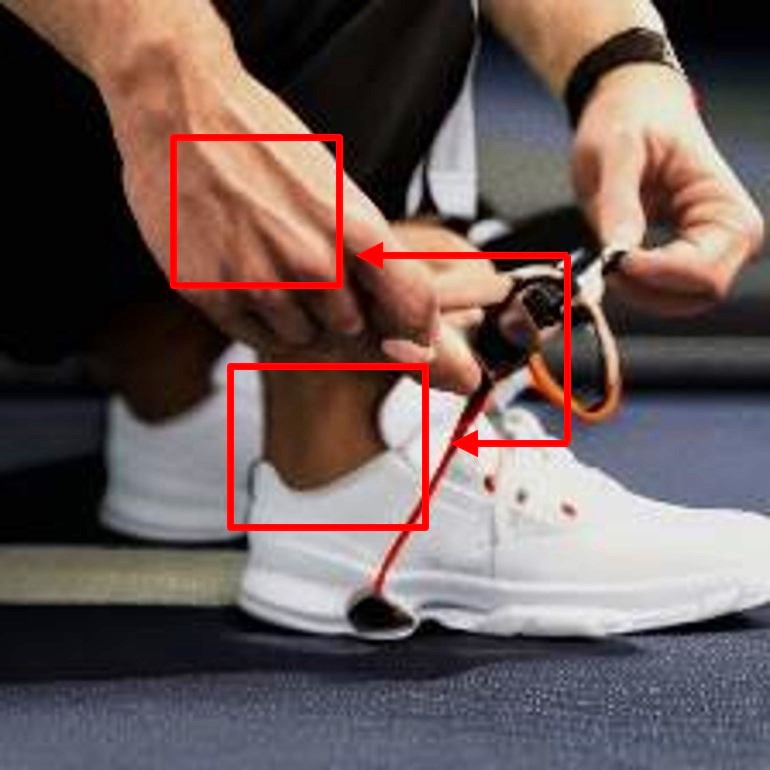,width=2cm}}
  \centerline{(d)  Color}\medskip

\end{minipage}
\begin{minipage}[b]{0.24\linewidth}
  \centerline{\epsfig{figure=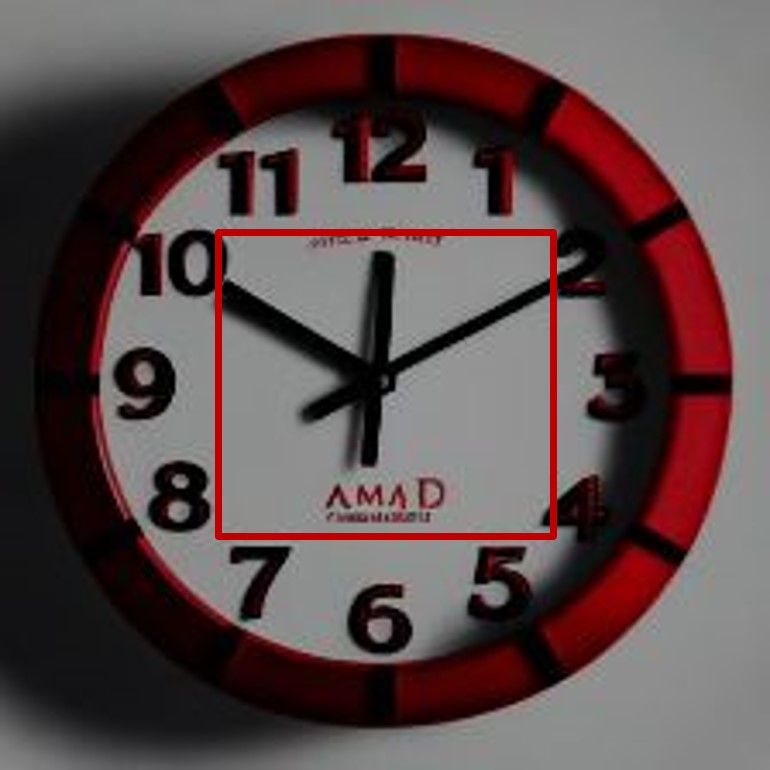,width=2cm}}
  
\end{minipage}
\hfill
\begin{minipage}[b]{.24\linewidth}
  \centerline{\epsfig{figure=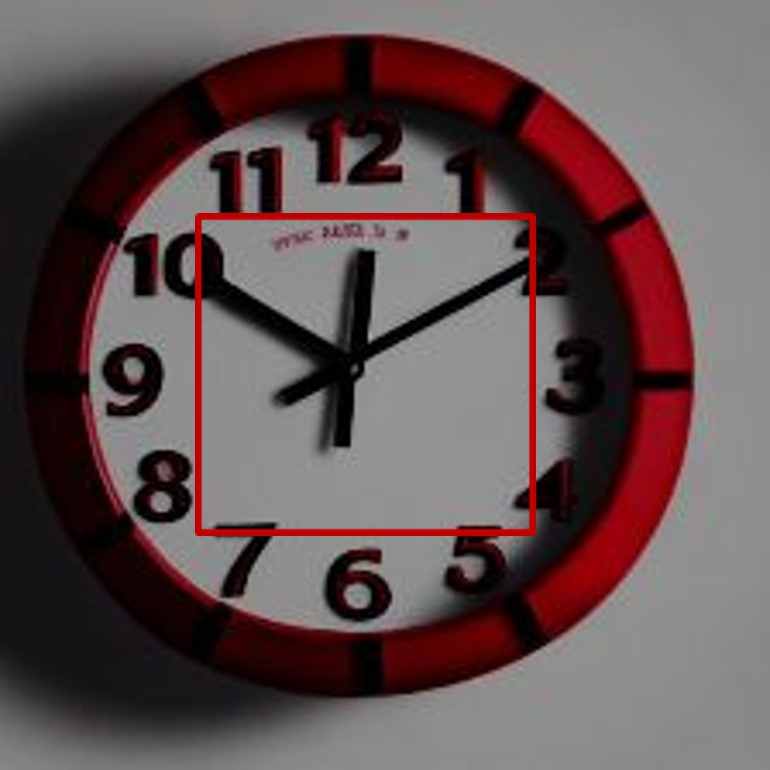,width=2cm}}
\end{minipage}
\hfill
\begin{minipage}[b]{0.24\linewidth}
  \centerline{\epsfig{figure=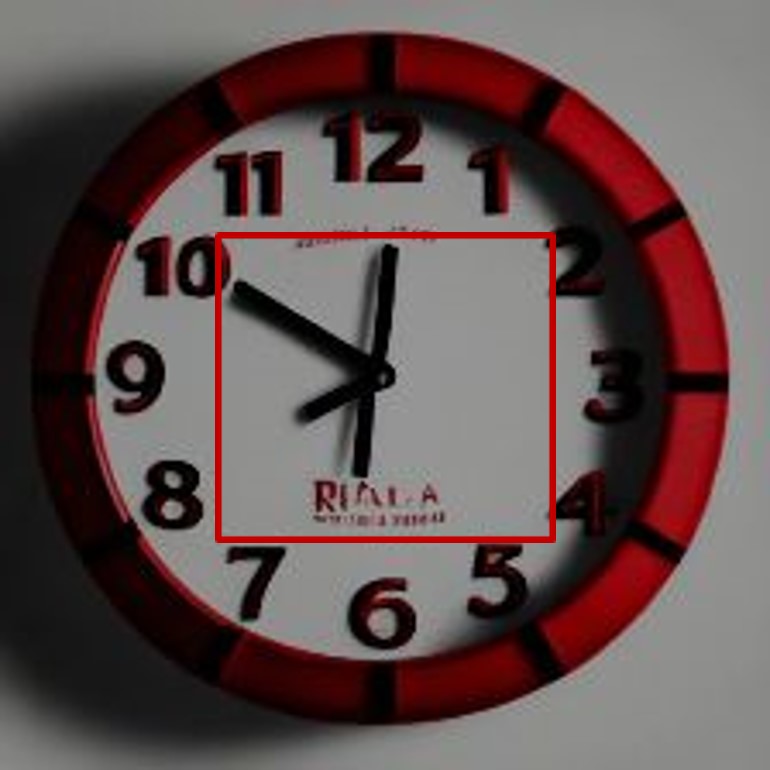,width=2cm}}
\end{minipage}
\hfill
\begin{minipage}[b]{0.24\linewidth}
  \centerline{\epsfig{figure=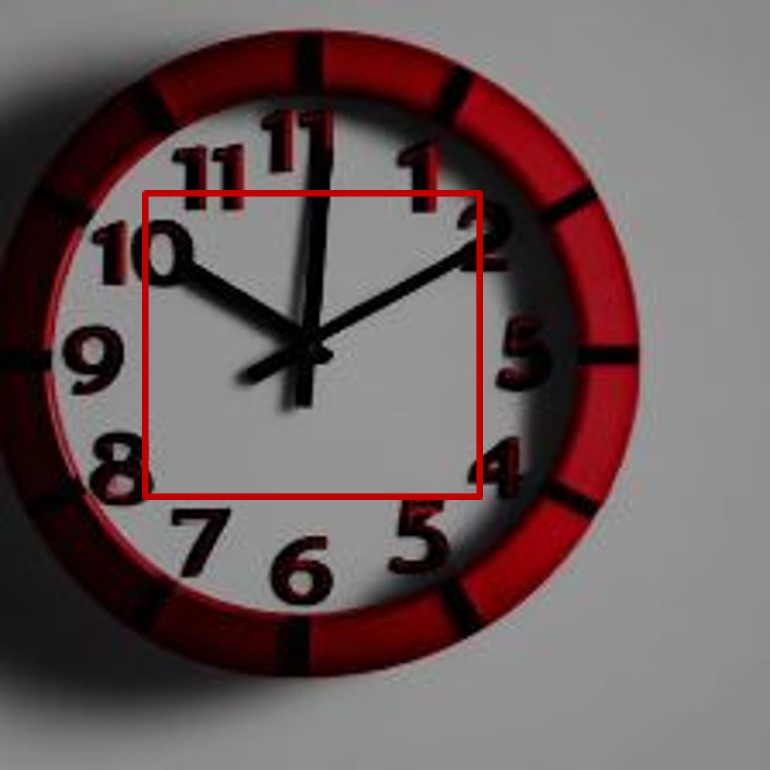,width=2cm}}
\end{minipage}
\centerline{(e) Temporal}\medskip
\caption{Illustration of spatial and temporal artifacts on AI-generated video. Spatial artifacts: (a) errors in geometric appearance, (b) errors in image layout, (c) frequency inconsistency, (d) color mismatch; Temporal artifacts: (e) mismatch between frames.}
\label{fig:1}

\end{figure}

\section{Dataset Construction}
\label{sec:dataset}
To benchmark AI-generated video detectors, we have created the first dataset for the AI-generated video detection task: the $Generated$ $Video$ $Forensics$ (GVF) dataset. 
The GVF dataset consists of 964 triples and can be represented as $\mathcal{D}=\{(r, p, F)^{(i)}\}_{i=1}^{964}$, where $r$ and $p$ represent the real video and the corresponding text prompt, and $F$ represents 
the videos generated by the generation models using the text prompt $p$. In order to examine the generalizability of the detectors, we have selected four open-source and four commercial generation models employing different generation methods and model frameworks as generation models for negative samples. 

\begin{figure}[t]
\centerline{\includegraphics[width=0.8\columnwidth]{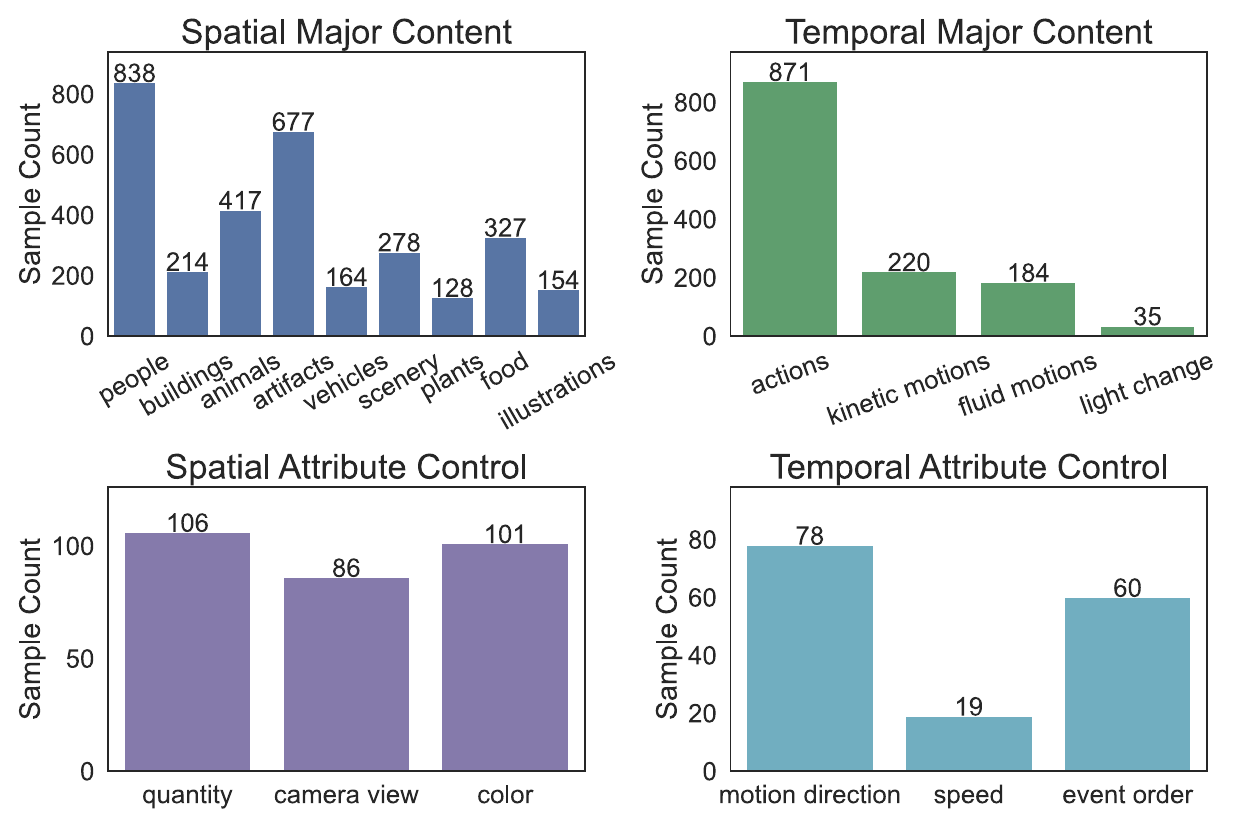}}
\caption{Data distribution over categories under the “major
content” (upper) and “attribute control” (lower) aspects.}
\label{fig:th4}
\end{figure}

\textbf{Prompt.} To approximate the distribution of videos in reality,
we have collected 964 prompts and corresponding real videos from open-domain text-to-video datasets: MSVD~\cite{chen2011collecting} and MSR-VTT~\cite{xu2016msr}. 
Following FTEV~\cite{liu2023fetv}, we consider the collection of prompts from the following four aspects to simulate the sample distribution in reality, rather than simply stacking the number of prompts: 
the spatial and temporal content it mainly describes and the spatial and temporal attributes it controls during video generation. 
Spatial content is classified as people, animals, plants, food, vehicles, buildings, and artifacts, with corresponding spatial attributes of quantity, color, and camera view. 
Temporal content encompasses actions, kinetic motions,  fluid motions, and light change, with attributes including motion direction, speed, and event order. 
This ensures that our GVF dataset is comprehensive and balanced.
Figure~\ref{fig:th4} illustrates the data distribution across categories under different aspects, which indicates that the GVF dataset contains enough data to cover different category ranges. \textit{Please refer to the supplementary materials for the complete construction process and detailed evaluation of the dataset.}



\section{Methodology}
\label{methodology}
\subsection{Probe experiments of STNN models}
\label{stcnn}
Initially, we considered using spatiotemporal model (STNN)~\cite{carreira2017quo,feichtenhofer2019slowfast,feichtenhofer2020x3d,fan2021multiscale}, which are widely used in video classification tasks, to train the generalizable AI-generated video detector. However, we discovered that these models are only effective on the specific generation models they were trained on, failing to discern unseen T2V generation models.
\begin{table}[h]
  \centering
  \caption{Spatiotemporal Neural Networks (STNN) rely on spatial artifacts.}

    \begin{tabular}{ccccccc}
    \toprule
    \multicolumn{1}{c}{\multirow{3}[6]{*}{Method}} & \multicolumn{4}{c}{Artifacts} & \multicolumn{2}{c}{\multirow{2}[4]{*}{Difference}} \\
\cmidrule{2-5}          & \multicolumn{2}{c}{  Temporal} & \multicolumn{2}{c}{Spatial} & \multicolumn{2}{c}{} \\
\cmidrule{2-7}          & ACC   & AUC    & ACC   & AUC    & ACC   & AUC \\
    \midrule
    I3D ~\cite{carreira2017quo}  & 51.17  & 51.61  & 86.67  & 98.61  & 35.50  & 47.00  \\
    Slow~\cite{feichtenhofer2019slowfast}  & 50.04  & 50.37  & 93.81  & 99.51  & 43.77  & 49.14  \\
    X3D~\cite{feichtenhofer2020x3d}   & 50.04  & 54.31  & 78.39  & 92.84  & 28.35  & 38.53  \\
    Mvit ~\cite{fan2021multiscale} & 50.48  & 55.29  & 72.16  & 92.59  & 21.68  & 37.30  \\
    \bottomrule
    \end{tabular}
  \label{tab:3DCNN_probing}%

\end{table}%
To analyze what information the STNN was learning during the training process, or rather, which features within the video it utilized to distinguish between authentic and fabricated content, we designed a probing experiment with two novel datasets under the same model framework. 
The first dataset involves scrambling the frames of all real videos in the current test set, destroying the temporal continuity of the videos, to observe if the model can differentiate based solely on the temporal discontinuities of the videos. The second dataset is created by randomly selecting a frame of fake videos from the current test set and replicating it multiple times to observe if the model could differentiate based on the spatial artifacts of the video frame alone.

The experiment results in Table~\ref{tab:3DCNN_probing} show that STNN overly relies on spatial artifacts, degrading the AI-generated video detection into a 2D problem. However, the fact that STNN relies on spatial artifacts still does not explain its lack of generalizability to unseen generation models. 
To further explore the possibility of detecting AI-generated video based on spatial artifacts, we turned to the SoTA AI-generated image detection works.
\subsection{Capturing Universal Spatial Artifacts is Difficult}
\label{short}

We transform the current detection of generated images into image-level detectors for detecting AI-generated videos and test them on the GVF dataset.
As shown in Table~\ref{tab:main}, these detectors demonstrate superior capabilities in capturing the spatial artifacts of specific generation models. However, they also lack generalizability to unseen generation models. 

To find out why these detectors lack generalizability to unseen generate models, we visualized the feature space of the detector~\cite{wang2020cnn} and examined the distribution of five sub-datasets: 
(\romannumeral1 ) Real, comprising real videos from the test set, (\romannumeral2) T2V, comprising fake videos from the sub-dataset Text2Video-zero,  (\romannumeral3) Show1, comprising fake videos from the sub-dataset Show-1, and (\romannumeral4) MP, comprising fake videos from the sub-dataset ModelScope, (\romannumeral5) ZP, comprising fake videos from the sub-dataset ZeroScope. 
\begin{figure}[h] 
\centering
\includegraphics[width=0.8\columnwidth]{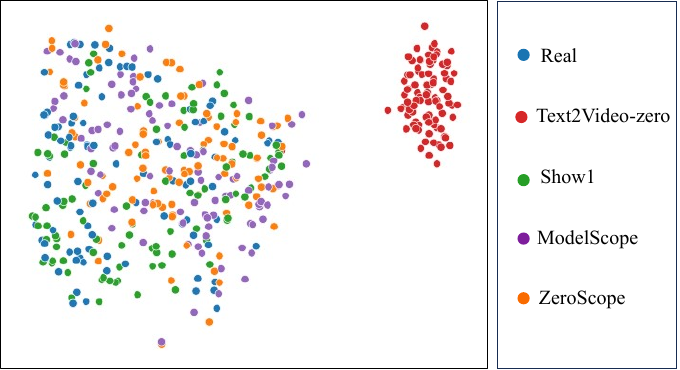}
\caption{t-SNE visualization of real and fake video frames associated with four video generation models.}
\label{fig_res50_stne}
\end{figure}
\begin{figure*}[t]
 \centerline{\includegraphics[width=0.8\textwidth]{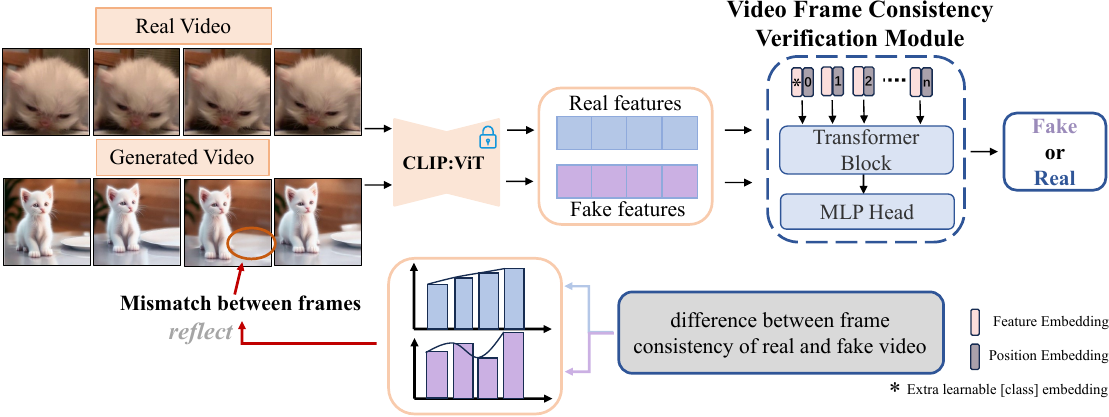}}
\caption{Overview of the DeCoF framework. We first get real video and AI-generated video features using the pre-trained CLIP:VIT, to eliminate the impact of spatial artifacts on capturing temporal artifacts. Then a verification module consisting of two transformer layers and one MLP head is used to learn the differences between frame consistency of the real and fake videos.}
\label{fig:res}
\end{figure*}
\begin{figure}[h] 
\centering
\scalebox{0.8}{
         \begin{minipage}{\linewidth}
		
            \begin{minipage}[b]{0.18\linewidth}
              \centerline{\epsfig{figure=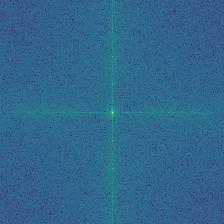,width=1.65cm}}
              \centerline{Real}\medskip
              
            \end{minipage}
            \hfill
            \begin{minipage}[b]{.18\linewidth}
              \centerline{\epsfig{figure=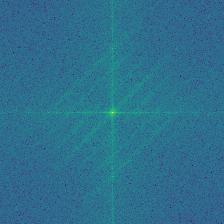,width=1.65cm}}
              \centerline{T2V}\medskip
            \end{minipage}
            \hfill
            \begin{minipage}[b]{0.18\linewidth}
              \centerline{\epsfig{figure=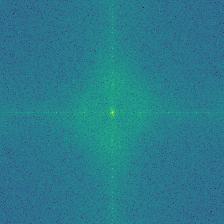,width=1.65cm}}
              \centerline{Show1}\medskip
            \end{minipage}
            \hfill
            \begin{minipage}[b]{0.18\linewidth}
              \centerline{\epsfig{figure=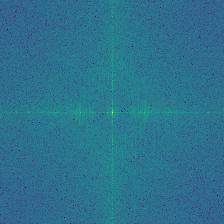,width=1.65cm}}
             \centerline{MP}\medskip
            
            \end{minipage}
            \hfill
            \begin{minipage}[b]{0.18\linewidth}
              \centerline{\epsfig{figure=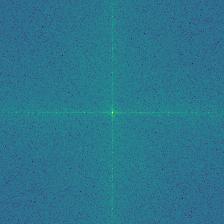,width=1.65cm}}
              \centerline{ZP}\medskip
            \end{minipage}
	   \end{minipage}}
\caption{Average spectrum of video frames for real videos and fake videos generated by different video generation models.}
            \label{fig:FFT}
\end{figure}

For each video, we utilized the detector trained on (\romannumeral1) and (\romannumeral2 ) to extract their feature representations corresponding to each frame and subsequently visualized these through t-SNE~\cite{van2008visualizing}.
As shown in Figure~\ref{fig_res50_stne}, on the one hand, the detector has successfully classified the real video and the fake video generated by the seen generate model (Text2Video-zero), indicating that the detector has fully captured the spatial artifacts of the videos generated by the seen generate model. On the other hand, the detector cannot distinguish fake videos generated by unseen models from real videos. Why are the spatial artifacts captured by the detector in (\romannumeral2) not helpful for detecting videos generated by other video generation models? We illustrate this by visualizing the frequency spectrum of video frames generated by the models. As shown in Figure~\ref{fig:FFT}, although the AI-generated videos all have spatial artifacts, such as parallel slashes appearing in T2V and bright spots appearing in Show-1, these spatial artifacts are not consistent. Therefore, capturing universal spatial artifacts appears to be extremely difficult.
\subsection{Detection based on Temporal Artifacts}
\label{method}
The recent video generation models convert the T2I model into the T2V model through additional constraints, which have common drawbacks in controlling video continuity in two aspects: 1) the normativity of motion, not only do visible movements violate to physical laws, but more importantly, there are irregular movements of invisible detail pixels, 2) there is a slight pixel shift in the relatively static local background. Therefore, we propose to develop a detection method based on temporal artifacts.

Given a set of video frames $F=\{F_{1}, F_{2}, \cdots, F_{L} \}$, we define the method to determine the authenticity of a video based on its temporal artifacts as $f(F_{1}, F_{2}, \cdots, F_{L})=y$, where $f$ is a specific function that measures the consistency between video frames, $y \in Y=\{0,1\}$ is the authenticity label of the video. It appears that the detection model tends to learn spatial artifacts (Section~\ref{stcnn}). 
\textit{Therefore, we propose to eliminate the impact of spatial artifacts in order to help the model focus on capturing temporal artifacts.}

Intuitively, we want to find a particular mapping function $M$ with the following properties: the distance between mapping frames is negatively related to the image similarity, and the mapping function is insensitive to spatial artifacts.
Hence, we choose ViT-L/14~\cite{dosovitskiy2020image}  trained for the task of image-language alignment as mapping function $M$. Then by mapping the video frames $F_{i}$ to $M(F_{i})$, We have implemented decoupling the temporal and spatial artifacts because the mapping function is not trained to capture spatial artifacts, but only encodes the image into a feature space with excellent properties, where the distance between features is inversely proportional to the image similarity.
As a result, the original problem can be expressed as $f(M(F_{1}), M(F_{2}), \cdots, M(F_{L}))=y$.
\begin{table*}[t]
  \centering
     \caption{Comprehensive comparison of our method with previous detectors and STNN, including CNNDet, DIF, DIRE, Lgard, I3D, Slow, X3D, and Mvit, were tested on our GVF dataset using the officially provided model. * Denotes we retrained it on a subdataset of GVF dataset with official code. We report ACC (\%) and AUC (\%) (ACC and AUC in the Table).}
    \scalebox{0.97}{
    \centering
    \begin{tabular}{ccccccccccccrr}
    \toprule
    \multirow{3}[6]{*}{Method } & \multicolumn{1}{c}{\multirow{3}[6]{*}{\makecell[c]{Train\\dataset}}} & \multicolumn{1}{c}{\multirow{3}[6]{*}{\makecell[c]{Generation\\mode}}} & \multicolumn{1}{c}{\multirow{3}[6]{*}{\makecell[c]{Detection\\Level}}} & \multicolumn{8}{c}{Video generation models}              & \multicolumn{2}{c}{\multirow{2}[4]{*}{\makecell[c]{Total\\Avg.}}} \\
\cmidrule{5-12}          &       &       &       & \multicolumn{2}{c}{T2V-Zero} & \multicolumn{2}{c}{Show1} & \multicolumn{2}{c}{ModelScope} & \multicolumn{2}{c}{ZeroScope} & \multicolumn{2}{c}{} \\
\cmidrule{5-14}          &       &       &       & ACC   & AUC    & ACC   & AUC    & ACC   & AUC    & ACC   & AUC    & \multicolumn{1}{c}{ACC} & \multicolumn{1}{c}{AUC} \\
    \midrule
    CNNDet(0.5)~\cite{wang2020cnn} &  - & ProGAN & Image & 49.61  & 41.47  & 49.74  & 47.58  & 49.61  & 43.40  & 49.94  & 59.85  & 49.73  & 48.08  \\
    CNNDet(0.1))~\cite{wang2020cnn} &  - & ProGAN & Image & 49.93  & 45.72  & 50.06  & 46.26  & 50.19  & 38.25  & 50.06  & 54.40  & 50.06  & 46.15  \\
    DIF~\cite{sinitsa2023deep}   & -  & Diffusion & Image & 35.10  & 47.96  & 45.90  & 53.42  & 31.40  & 46.91  & 27.50  & 46.59  & 34.98  & 48.72  \\
    DIRE~\cite{wang2023dire}  & - & Diffusion & Image & 63.14  & 63.45  & 51.80  & 51.70  & 51.48  & 51.03  & 46.71  & 42.96  & 53.28  & 52.29  \\
    Lgard~\cite{tan2023learning} & - & ProGAN & Image & 53.09  & 64.78  & 59.70  & 68.42  & 40.21  & 38.87  & 47.94  & 51.80  & 50.24  & 55.97  \\
     UFDetect~\cite{ojha2023towards} & - &ProGAN	&Image	&50.13 &63.99	&53.09 &74.42	&52.13 &75.55	&58.89 &84.27	&53.56 &74.56\\
    CNNDet (0.5)*)~\cite{wang2020cnn}   & T2V-Zero& T2V& Image & 98.64  & 99.97  & 51.93  & 63.12  & 51.54  & 71.07  & 54.96  & 72.43  & 64.27  & 76.64 \\
    CNNDet (0.1)*)~\cite{wang2020cnn}  & T2V-Zero& T2V & Image & 96.84  & 99.99  & 53.93  & 67.10  & 55.34  & 73.67  & 60.69  & 75.76  & 66.70  & 79.13  \\
    DIRE*~\cite{wang2023dire}  & T2V-Zero& T2V & Image & 99.17  & 99.99  & 49.93  & 56.74  & 52.83  & 77.58  & 51.28  & 67.12  & 63.30  & 75.36  \\
   Lgard*~\cite{tan2023learning}    & T2V-Zero& T2V & Image & 98.45  & 100.00  & 56.19  & 78.91  & 51.03  & 66.30  & 54.64  & 75.93  & 65.08  & 80.29  \\
   UFDetect*~\cite{ojha2023towards}  &T2V-Zero	& T2V&Image	&99.36 & 99.77	&62.31 &88.9	&67.14 &90.70	&57.73 &86.37	&73.21 &91.44\\
    \midrule
    I3D*~\cite{carreira2017quo}    & T2V-Zero& T2V & Video & 98.00  & 99.87  & 49.75  & 70.41  & 49.25  & 71.43  & 50.50  & \multicolumn{1}{r}{51.85 } & 61.88  & 73.39  \\
    Slow*~\cite{feichtenhofer2019slowfast}  & T2V-Zero& T2V & Video & 93.30  & 99.48  & 50.48  & 52.09  & 50.59  & 49.43  & 49.41  & 37.29  & 60.95  & 59.57  \\
    X3D*~\cite{feichtenhofer2020x3d}   & T2V-Zero& T2V & Video & 94.84  & 98.61  & 49.41  & 33.52  & 48.35  & 43.07  & 49.85  & 30.89  & 60.61  & 51.52  \\
    Mvit*~\cite{fan2021multiscale} & T2V-Zero& T2V & Video & 98.42  & 99.97  & 49.96  & 45.93  & 49.49  & 38.98  & 51.68  & 32.60  & 62.39  & 54.37  \\
    EVR*~\cite{lin2022frozen}	&T2V-Zero & T2V	&Video	&97.50 &99.93	&78.5 &90.96	&59.75 &77.97	&64.75 &85.46	&67.67 &88.58 \\
    \midrule
    DeCoF   & T2V-Zero& T2V & Video & 99.50  & 100.00  & 71.50  & 95.83  & 78.25  & 96.13  & 76.50  & 96.13  & 81.44 & 97.02  \\
    DeCoF  & Show1   & T2V  & Video & 98.50  & 99.77  & 97.00  & 99.33  & 90.25  & 97.96  & 93.75  & 99.08  & \textbf{94.88}  & \textbf{99.04}  \\
    DeCoF  & ModelScope   & T2V  & Video &  98.50  & 99.94  & 80.25  & 95.68  & 97.00  & 99.53  & 93.75  & 98.93  & 92.38  & 98.52  \\
    DeCoF  &  ZeroScope   &  T2V  & Video & 97.00  & 99.81  & 75.50  & 96.52  & 75.75  & 96.42  & 98.50  & 99.91  & 86.69  & 98.17  \\
    \bottomrule
    \end{tabular}%
    }
  \label{tab:main}%

\end{table*}%

\textbf{Model framework.} In order to learn video frame consistency for capturing temporal artifacts, we design a \textbf{de}tection method based on the \textbf{Co}nsistency of video \textbf{F}rames (DeCoF), in which the core verification module consists of two transformer layers and an MLP head as shown in Figure~\ref{fig:res}.
More specifically, given a video input $V$, the frozen parameter image encoder (ViT-L/14) extracts the features of the video frames and generates a feature sequence $S=\{M(F_{1}), M(F_{2}), \cdots, M(F_{L})\}$ with a shape of $L*D$, where $D$ represents the dimension of each feature. Next, the position encoding and a learnable class embedding representing the classification output are prepended to the sequence $S$ and then passed through a sequential module composed of two layers of transformers.
Finally, a classification header is appended to the transformer layer, outputting a probability distribution over the target class. We train our model via a simple cross-entropy loss.

\section{Experiments}
\begin{table*}[ht]
     \caption{Performance of DeCoF on commercial generation models.}
     \centering
    \scalebox{1}{
    
    \begin{tabular}{ccccccccccccc}
    \toprule
    \multicolumn{1}{c}{\multirow{3}[6]{*}{\makecell[c]{Train\\dataset}}}
    & \multicolumn{10}{c}{Video generation models} & \multicolumn{2}{c}{\multirow{2}[4]{*}{\makecell[c]{Total\\Avg.}}}\\
\cmidrule{2-11}     & \multicolumn{2}{c}{Gen-2} & \multicolumn{2}{c}{Pika} & \multicolumn{2}{c}{Sora} & \multicolumn{2}{c}{Veo} &\multicolumn{2}{c}{Kling} & \multicolumn{2}{c}{}\\
\cmidrule{2-13}          & ACC   & AUC    & ACC   & AUC    & ACC   & AUC  & ACC   & AUC & ACC   & AUC & ACC   & AUC\\
    \midrule
    \multicolumn{1}{c}{Text2Video-zero} & 99.49  & 100.00  & 99.49  & 99.98  & 96.87  & 	99.70  &96.43 &100.00	&89.89&98.91&96.43&99.72 \\
    \multicolumn{1}{c}{Show1} & 98.46  & 99.96  & 98.46  & 99.85  & 94.79  & 	98.26  &92.86 &100.00 &89.36&98.45&94.79&99.30\\
    ModelScope & 98.97  & 99.99  & 96.89  & 99.73  & 94.79  & 	 98.83	&89.29 &98.47&95.21&99.14&95.03&99.23 \\
    ZeroScope & 98.97  & 99.99  & 94.84  & 99.52  & 89.58  & 	97.92	&67.86 &94.90&78.72&98.10&85.87&98.10 \\
    \bottomrule
    \end{tabular}%
    }

  \label{tab:gen2}%
 
\end{table*}%

\subsection{Experimental Setup} 
\label{set}

\textbf{Data pre-processing.} We only train on a certain sub-dataset of GVF and test on the remaining sub-datasets to evaluate the generalizability of the detector. 
We uniformly sample 8 frames from the first 32 frames of each video and center-crop each frame according to its shortest side, then resize it to 224 × 224. During the training process, some data augmentation, such as Jpeg compression, random horizontal flipping, and Gaussian blur are also used. 

\textbf{Evaluation metrics.}
Following previous methods~\cite{wang2020cnn,wang2023dire,frank2020leveraging}, we mainly report video-level accuracy (ACC) and area under
curve (AUC) in our experiments to evaluate the detectors. The threshold for computing accuracy is set to 0.5 following~\cite{wang2020cnn,wang2023dire}. For image-based methods, we average frame-level predictions into corresponding video-level predictions.
\subsection{Evaluation }
We conduct comprehensive evaluations to answer the following questions:

\textbf{Q1. How does our detector generalize to unseen T2V models compared to previous detectors?}

Table~\ref{tab:main} compares CNNDet~\cite{wang2020cnn}, DIF~\cite{sinitsa2023deep}, DIRE~\cite{wang2023dire}, Lgard~\cite{tan2023learning}, UFDetect~\cite{ojha2023towards}, using the pre-trained weights provided by the official repository. Although these detectors have good performance in detecting generated images, their performance drops significantly on the task of detect AI-generated videos, with both ACC and AUC below 60\%.
As a result of retraining the above five detectors on the sub-dataset Text2Video-zero, their performance on seen video generation models greatly improved, but they still performed poorly for unseen generation models, with ACC and AUC scores still lower than 60\%, and the retrained STNN presents similar results.
On the contrary, our DeCoF has excellent generalizability for various unseen generation models. Compared with previous video classification methods based on pre-trained CLIP~\cite{lin2022frozen}, our method has still made significant progress. Specifically, on the task of AI-generated video detection,  DeCoF outperforms previous detectors by 8.23\% $ \sim $  31.38\% in ACC and  5.58\% $\sim$  49.18\% in AUC. 
We also conducted cross-dataset evaluations. As shown in Table~\ref{tab:main}, the ACC of the model trained on the other three data sets is much higher than that of the model trained on Text2Video-zero by 5.25\% $ \sim $ 13.44\%.

\textbf{Q2. How does our detector maintain robustness against noise and disturbances?}

In practical detection scenarios, the robustness of the detector to unseen perturbations is also crucial. Here, we mainly focus on the impact of two perturbations on the detector: Gaussian blur and JPEG compression. The perturbations are added under three levels for Gaussian blur ($\sigma$ = 1, 2, 3) and five levels for JPEG compression (quality = 90, 80, 70, 60, 50). Figure~\ref{fig:robu} shows the performance of our model trained on different sub-datasets in the face of the above two perturbations. Although experiencing some performance inconsistencies due to differences between training datasets, our method remains effective under all degrees of arbitrary perturbation, with the AUC  higher than 80\%. \\

\section{Conclusions and Future Work}
The paper presents the first investigation on AI-generated video detection, with the goal of tackling the societal concern of validating the authenticity of video content.
We construct the first public dataset for video detection, providing a solid benchmark for community research. 
At the same time, we propose a simple yet effective detection model, and prove its strong generalizability and robustness on multiple open-source and commercial closed-source models.
We hope this study will inspire the creation and improvement of other detection technologies, providing new avenues for the development of authentic and reliable AIGC applications.
\bibliographystyle{IEEEbib}
\bibliography{icme2025references}

\clearpage
\appendix
\section{Appendix}
This supplementary material provides:
\begin{itemize}
    \item Sec.~\ref{A}: A detailed description of dataset construction, including dataset structure diagrams, as well as specific examples of prompts and videos.
    \item Sec.~\ref{B}: Various parameter details for model training, hyperparameter settings, etc.
     \item Sec.~\ref{K}: Robustness in the face of unknown disturbances.
    \item Sec.~\ref{C}: The impact of selecting different mapping functions.
    \item Sec.~\ref{D}: The impact of video frame count.
    \item Sec.~\ref{E}: The impact of video frame sampling interval.
    \item Sec.~\ref{F}: More details on Sora and Veo.
    \item Sec.~\ref{G} : Spatial artifacts vs Temporal artifacts.
    \item Sec.~\ref{H} : Impact Statements.
    \item Sec.~\ref{I} : Limitation.
    \item Sec.~\ref{J}: More visual examples, Including examples of generated videos and real videos, as well as Grad CAM visualization of our method on different generated videos.

\end{itemize}
\subsection{$GeneratedVideoForensics$ Dataset}
\label{A}
\subsubsection{Problem Formulation}
\textbf{Generated Video Detection.} In addition to the generalizability to unseen generation models, we also take the generalizability to cross-category videos into account. The generated video detection task can be defined as follows:  

Given a collection of real videos denoted as $R=\{r_{1}, r_{2}, \cdots,r_{n}\}$, a set of content prompts  $P=\{p_{1}, p_{2},\cdots,p_{n}\}$, 
where $r_{i}$ and $p_{i}$, $i \in [1, n]$ refer to each real video and its corresponding content prompt.
Let $G=\{g_{1}, g_{2},\cdots,g_{m} \}$ denote a set of video generation models. Using the prompt $p_{i}$ of each real video $r_{i}$, $G$ generates $m$ corresponding videos, resulting in a set of fake video collections $F=\{F_{1}, F_{2},\cdots, F_{m} \}$, where $F_{j}=\{g_{j}(p_{1}),  g_{j}(p_{2}), \cdots, g_{j}(p_{n}) \}$, $j \in [1, m]$.
As such, we form the generated video detection dataset $\mathcal{D}=\{D_{1}, D_{2},\cdots,D_{m}\}$,
where $D_{j}=\{(x,y)^{(i)}\}_{i=1}^{n}$, with $\mathcal{X}=\{R, F_{j}\}\subset \mathbb{R}^{L*H*W*C}$, $\mathcal{Y}=\{0,1\}$, $x\in\mathcal{X}$ and $y \in \mathcal{Y}$ denoting a video and its label, $L, H, W $ and $C $ correspond to the video length, video frame height, width and number of channels, respectively.

The goal is to train a detector $\mathcal{F}$ on a subdataset $D_{j}$ built based on a known generation model $g_{j}$, map the video space to the category space $\mathcal{F}$ $:$ $\mathcal{X}\rightarrow \mathcal{Y}$, and generalizes to unseen generation models $G_{k}$, where $k \in [1, m], k \neq j$.
This can be achieved by minimizing  classification error of $\mathcal{F}$  on training data $D^{*}_{i}$:
\begin{eqnarray}
\mathop{\arg \min} \limits_{\theta}  \mathbb{E}_{(x,y)\in D^{*}_{i}} \mathcal{L}(f(x),y)
\end{eqnarray}
where $\mathcal{L}$ is a loss function, and $\theta$ are the trainable parameters of detector $f$, $D^{*}_{i}$ refers to the training set of $D_{i}$.  We evaluate the detection of cross-category-generated videos by ensuring that the training set and the test set do not contain videos using the same prompt.
\subsubsection{Dataset Composition}

The following section provides a more detailed description of how the $GeneratedVideoForensics$ was constructed and evaluated. The entire GVF dataset is generated in two steps:

a. Selection of prompts and real videos: The prompts and real videos of the GVF dataset come from existing open-domain text-to-video datasets: MSVD ~\cite{chen2011collecting} and MSR-VTT ~\cite{xu2016msr} datasets. When selecting samples, we followed the work of FETV ~\cite{liu2023fetv}, Considering the collection of prompts from the above four aspects to simulate the sample distribution in real scenarios. More specifically, in the work of FETV, they classify each prompt  from two aspects: the main content it describes and the attributes it controls during video generation.
The spatial major content of the prompt can be divided into people, animals, plants, food, vehicles, buildings, artifacts, scenery, and illustrations. The corresponding spatial attribute controls are divided into quantity, color, and camera view. 
Likewise, the major contents and attribute controls of temporal are actions, kinetic motions, light change, and fluid motions, respectively, motion direction, speed, and event sequence. 

b. Selection of generation models: Although there are many high-quality generation models on the market, we do not want to choose the best. On the contrary, we have selected four generation models with their advantages, aiming to simulate the crises we are encountering in real life: a large number of generation models appear, how do we deal with the problems caused by unseen generation models? In the end, the four most representative generation models were selected as our generation models: \textbf{Text2Video-zero}~\cite{khachatryan2023text2video}, \textbf{ModelScopeT2V}~\cite{wang2023modelscope}, \textbf{ZeroScope}\cite{zero}, \textbf{Show1}~\cite{zhang2023show}. At the same time, we also selected the three most popular models as the secret test dataset: Pika~\cite{Pika}, Gen-2~\cite{Gen-2}, Kling~\cite{Kling}, Sora\cite{Sora} and Veo~\cite{Veo}. For Sora and Veo, we  provide detailed information in the section~\ref{F}

If divided by content, the $GenerateVideoForensics$ data set $D$ consists of 964 triples, which can be expressed as $D=\{(R, P, F)^{(i)}\}_{i=1}^{964}$, $R$ and $P$ represent real videos and corresponding text prompts, $F$ represents multiple fake videos generated by different generation models based on text prompts $P$. 

If divided by the generation model, the GenerateVideoForensics data set can be divided into four sub-datasets: Text2Video-zero, ModelScopeT2V, ZeroScope, and Show-1. Each sub-dataset consists of 964 new triples, which can be expressed as $D=\{(R, P, f)^{(i)}\}_{i=1}^{964}$, $R$ and $P$ represent the real video and the corresponding text prompt, and $f$ represents the fake video generated by the specific generation model represented by the name of the current sub-dataset based on the text prompt $P$. 

\begin{figure}[t]
\begin{minipage}[b]{0.05\linewidth}
  \centerline{\rotatebox{90}{T2V-Zero}}
  \vspace{0.5cm}
  \centerline{\rotatebox{90}{SHOW1}}
  \vspace{0.4cm}
  \centerline{\rotatebox{90}{MScope}}
  \vspace{0.4cm}
  \centerline{\rotatebox{90}{ZScope}}
  \vspace{0.8cm}
  \centerline{\rotatebox{90}{Real}}
  \vspace{0.4cm}
\end{minipage}
\begin{minipage}[b]{0.17\linewidth}
  \centering
  \centerline{Time 0}
  \centerline{\epsfig{figure=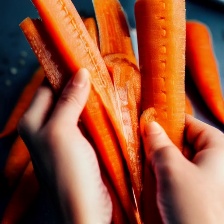,width=1.5cm}}
  \vspace{0.05cm}
  \centerline{\epsfig{figure=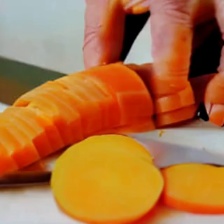,width=1.5cm}}
  \vspace{0.05cm}
  \centerline{\epsfig{figure=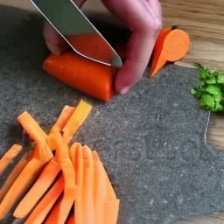,width=1.5cm}}
  \vspace{0.05cm}
  \centerline{\epsfig{figure=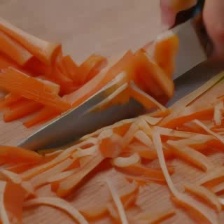,width=1.5cm}}
  \vspace{0.05cm}
  \centerline{\epsfig{figure=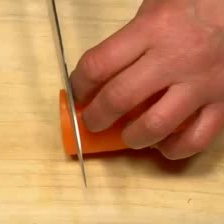,width=1.5cm}}

\end{minipage}
\hfill
\begin{minipage}[b]{0.17\linewidth}
  \centering
  \centerline{Time 1}
  \centerline{\epsfig{figure=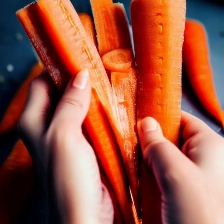,width=1.5cm}}
  \vspace{0.05cm}
  \centerline{\epsfig{figure=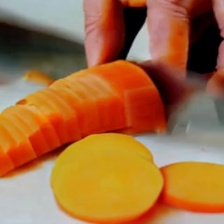,width=1.5cm}}
  \vspace{0.05cm}
  \centerline{\epsfig{figure=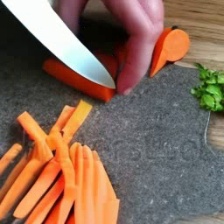,width=1.5cm}}
  \vspace{0.05cm}
  \centerline{\epsfig{figure=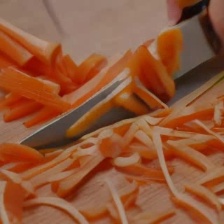,width=1.5cm}}
  \vspace{0.05cm}
  \centerline{\epsfig{figure=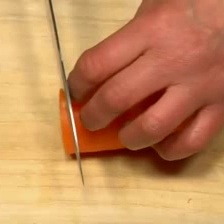,width=1.5cm}}

\end{minipage}
\hfill
\begin{minipage}[b]{0.17\linewidth}
  \centering
  \centerline{Time 2}
  \centerline{\epsfig{figure=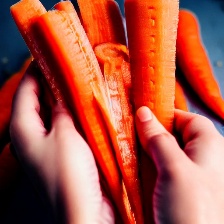,width=1.5cm}}
  \vspace{0.05cm}
  \centerline{\epsfig{figure=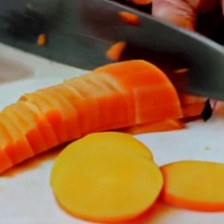,width=1.5cm}}
  \vspace{0.05cm}
  \centerline{\epsfig{figure=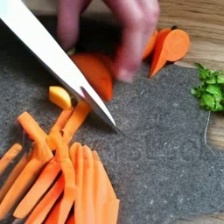,width=1.5cm}}
  \vspace{0.05cm}
  \centerline{\epsfig{figure=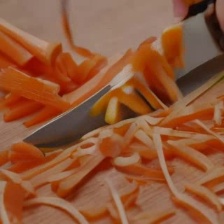,width=1.5cm}}
  \vspace{0.05cm}
  \centerline{\epsfig{figure=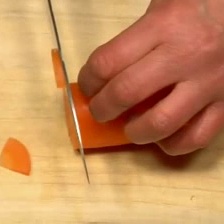,width=1.5cm}}
 
\end{minipage}
\hfill
\begin{minipage}[b]{0.17\linewidth}
  \centering
  \centerline{Time 3}
  \centerline{\epsfig{figure=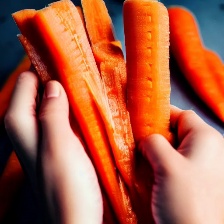,width=1.5cm}}
  \vspace{0.05cm}
  \centerline{\epsfig{figure=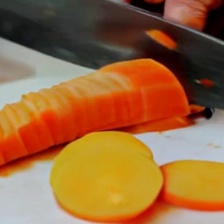,width=1.5cm}}
  \vspace{0.05cm}
  \centerline{\epsfig{figure=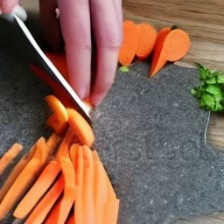,width=1.5cm}}
  \vspace{0.05cm}
  \centerline{\epsfig{figure=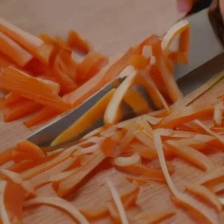,width=1.5cm}}
  \vspace{0.05cm}
  \centerline{\epsfig{figure=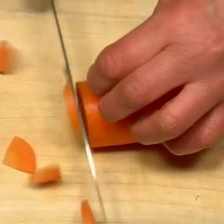,width=1.5cm}} 
\end{minipage}
\hfill
\begin{minipage}[b]{0.17\linewidth}
  \centering
  \centerline{Time 4}
  \centerline{\epsfig{figure=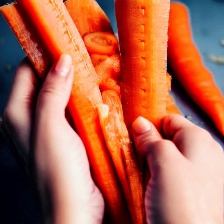,width=1.5cm}}
  \vspace{0.05cm}
  \centerline{\epsfig{figure=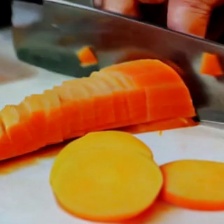,width=1.5cm}}
  \vspace{0.05cm}
  \centerline{\epsfig{figure=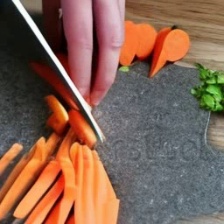,width=1.5cm}}
  \vspace{0.05cm}
  \centerline{\epsfig{figure=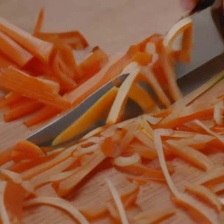,width=1.5cm}}
  \vspace{0.05cm}
  \centerline{\epsfig{figure=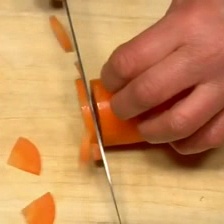,width=1.5cm}}
\end{minipage}
\caption {We generated four sub-datasets based on the real videos using the prompts and selected generation models employing four different architectures and generation methods: Text2Video-zero, ModelScopeT2V, ZeroScope, and Show-1. Each pair of positive and negative samples use the same prompt. Here shows the first five frames of four sample pairs using the prompt: \textit{A person is cutting a carrot}.}
\label{fig:2}
\end{figure}

\subsubsection{Dataset Evaluation}

As part of this research, we present a comprehensive analysis of the GVF dataset, which includes an analysis of the attribute statistics of prompts, as well as an assessment of the different generation models in terms of the spatial content, spatial attributes, temporal content, and temporal attributes according to the FETV~\cite{liu2023fetv}.

\begin{table*}[htbp]
  \centering
  \caption{\textbf{Some examples of prompts.} Each prompt is composed of a random combination of four attributes: main content of time, main attributes of time, main content of space, and main attributes of space. Here we list some examples of prompts, \textcolor{red}{red }for content and \textcolor{blue}{blue} for attributes.}
    \begin{tabular}{c|cc|cc|cccccc}
    \toprule
          & \multicolumn{2}{c|}{Main content} & \multicolumn{2}{c|}{Attribute \newline{}control} & \multicolumn{6}{c}{Example} \\
    \midrule
    \multirow{4}[2]{*}{Spatial} & \multicolumn{2}{c|}{People} & \multicolumn{2}{c|}{ Quantity} & \multicolumn{6}{c}{\textcolor{blue}{four} \textcolor{red}{friends} are driving in the car} \\
          & \multicolumn{2}{c|}{Animals} & \multicolumn{2}{c|}{Color} & \multicolumn{6}{c}{a \textcolor{blue}{white }\textcolor{red}{bird} chirps and plays on a handmade staircase} \\
          & \multicolumn{2}{c|}{Plants} & \multicolumn{2}{c|}{Camera View} & \multicolumn{6}{c}{a beautiful view of \textcolor{red}{greenery} and river from \textcolor{blue}{a top view}} \\
          & \multicolumn{2}{c|}{Food} & \multicolumn{2}{c|}{Color} & \multicolumn{6}{c}{in cup full of chocolate then we add \textcolor{blue}{yellow} \textcolor{red}{cream} to it so yummy} \\
    \midrule
    \multirow{4}[2]{*}{Temporal} & \multicolumn{2}{c|}{Actions} & \multicolumn{2}{c|}{Event Order} & \multicolumn{6}{c}{a red-headed woman is shown \textcolor{red}{walking} and \textcolor{blue}{then} \textcolor{red}{sitting}} \\
         & \multicolumn{2}{c|}{Actions} & \multicolumn{2}{c|}{Speed} & \multicolumn{6}{c}{a man \textcolor{blue}{fastly} \textcolor{red}{cross} the road by run and \textcolor{red}{get in} to the car} \\
          & \multicolumn{2}{c|}{Kinetic Motions} & \multicolumn{2}{c|}{Motion Direction} & \multicolumn{6}{c}{the big black  tractor is \textcolor{red}{moving} \textcolor{blue}{away} from it s carrier} \\
          & \multicolumn{2}{c|}{Fluid  Motions} & \multicolumn{2}{c|}{Speed} & \multicolumn{6}{c}{a bucket of slime is \textcolor{blue}{quickly} \textcolor{red}{poured} on top of a woman's head} \\
    \bottomrule
    \end{tabular}%

  \label{tab:prompts_exam}%
\end{table*}%

\begin{figure}[t]

  \centerline{\includegraphics[width=0.5\textwidth]{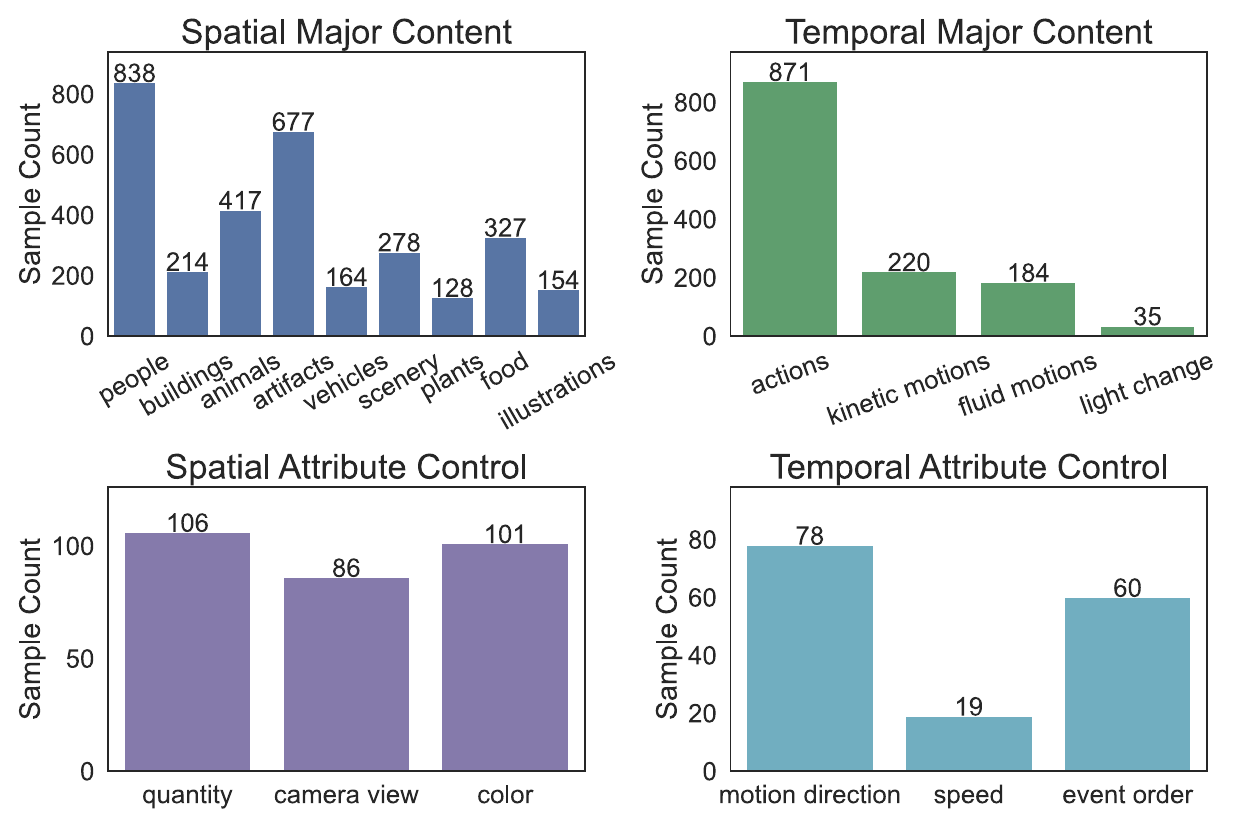}}
\caption{Data distribution over categories under the “major
content” (upper) and “attribute control” (lower) aspects.}
\label{fig:ap_s}
\end{figure}

\textbf{Attribute statistics of prompts.}
From the four aspects mentioned before, the main content of space and time and the main attributes of space and time, we provide attribute statistics of prompts of the data set. As the figure~\ref{fig:ap_s} shows, the various attributes of prompts are different because one prompt may have multiple attributes at the same time. At the same time, different attributes have different difficulties in generating videos, and the harm they cause in a trust crisis is also different. Here, we also provide some specific examples of prompts, As shown in the table~\ref{tab:prompts_exam}.

\textbf{Evaluation of different generation models.}
Here, we present our assessment of four generation models according to the FETV standard~\cite{liu2023fetv}. We assess the generated videos based on four perspectives: 
(\romannumeral1) Static quality focuses on the visual quality of single video frames. 
(\romannumeral2) Temporal quality focuses on the temporal coherence of video frames. 
(\romannumeral3) Overall alignment measures the overall alignment between a video and the given text prompt. 
(\romannumeral4) Fine-grained alignment measures the video-text alignment regarding specific attributes. 

\begin{figure}[h]
 \vskip -0.01in
  \centerline{\includegraphics[width=\columnwidth]{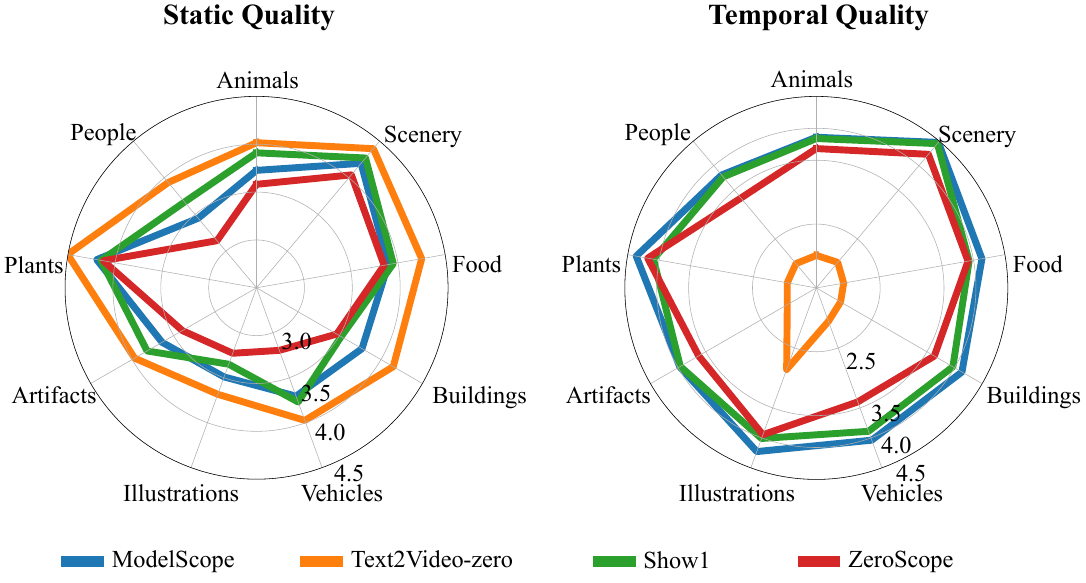}}
  \vskip 0.2in
 \centerline{\includegraphics[width=\columnwidth]{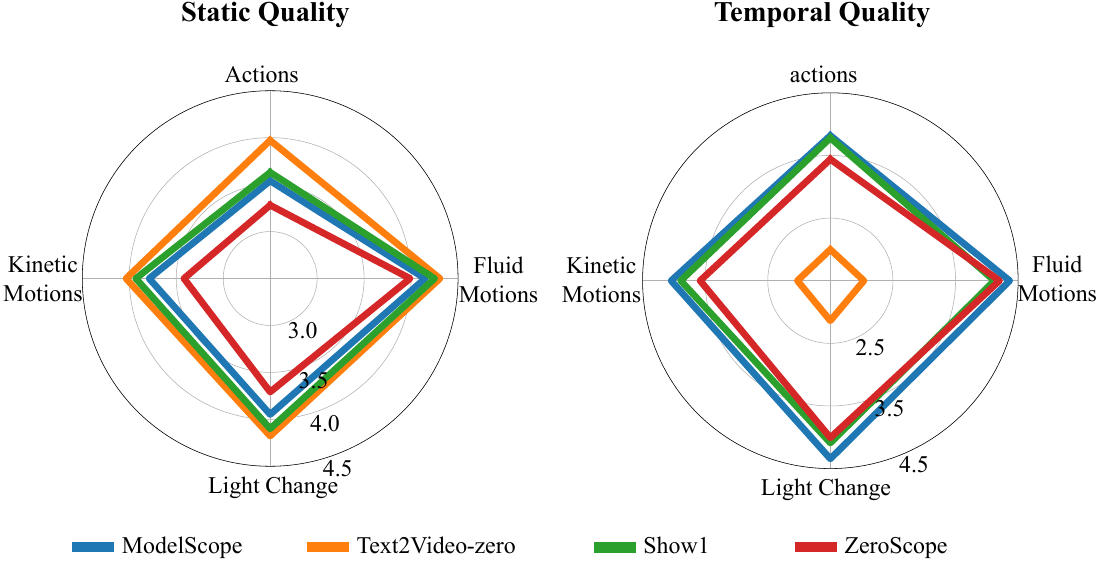}}
  \vskip 0.2in
\centerline{\includegraphics[width=\columnwidth]{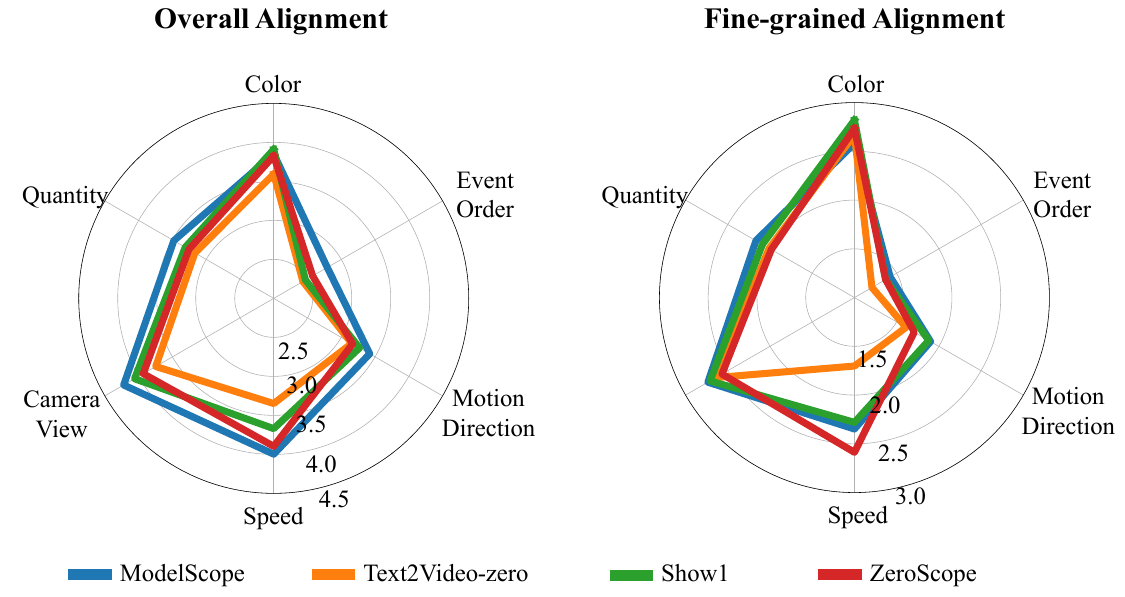}}
\caption{The manual evaluation provided results across major static and temporal contents and attributes to control video-text alignment.}
\label{fig:ap_th5}
\end{figure}
As shown in figure~\ref{fig:ap_th5}, Regarding spatial categories, the performance quality of these four data sets varies, and the performance of the same data set differs significantly between the static quality and temporal quality levels.
The videos that include people, animals, and vehicles have poor spatial quality, but the videos that include plants, scenery, and natural items have greater spatial quality. The reason behind this is that the videos in the first three categories contain numerous high-frequency intricacies, such as face and finger, which are more difficult to comprehend compared to the low-frequency information found in the final two categories, such as water and clouds.

In terms of static quality, Text2Video-zero performs well but performs poorly in terms of temporal quality. Text2Video-zero inherited the ability to generate images from Stable Diffusion, but it lacks temporal coherence in its videos due to the lack of video training. 
Show-1 is second only to Text2Video-zero in terms of static quality, and its temporal quality is also superior.
In comparison to ModelScopeT2V, ZeroScope is inferior to ModelScope in temporal quality but performs worse in static quality. 
The reason for this may be that ZeroScope has been primarily designed for videos with a 16:9 aspect ratio and without watermarks, rather than focusing on improving visual quality and aligning video text.

It is important to note that while the four models have better control over color and camera view, they do not perform as well when it comes to accurately controlling attributes such as quantity, motion direction, and order of the events.
In terms of speed and motion direction attribute control, ModelScope and ZeroScope are particularly promising, whereas Text2Video-zero demonstrates poor performance.

On different attributes of the four models, fine alignment and global alignment perform similarly, although there are some exceptions. This is due to the fact that other properties may have an effect on the overall alignment. Therefore, it is more accurate to assess controllability using fine alignment when examining a specific attribute, whereas global alignment is a reasonable approximation.

\subsection{Parameter Description}
\label{B}
In this section, we provide a detailed description of various parameters of the experiment, including the hyperparameters of the model and the details of its training.

\textbf{Training details.} During the training process, we only train on one sub-dataset of the GVF dataset and test on the other sub-datasets. The numbers of the training set, validation set, and test set are 771, 96, and 97, respectively. We freeze the parameters of CLIP:ViT~\cite{dosovitskiy2020image}, making it a pure mapping function, and only train our model using the SGD optimizer with an initial learning rate of 0.001 and a momentum coefficient of 0.9. All experiments were completed on two RTX 3090 graphics cards.

\textbf{Data pre-processing.}
All the following experiments are performed on our GVF dataset. In order to better detect the videos generated by the unseen generate model, we only train the model on the sub-dataset text2video-zero and test on the remaining three sub-datasets Show1~\cite{zhang2023show}, ModelScope~\cite{wang2023modelscope} and ZeroScope~\cite{zero}
Since the resolution and length of the videos generated by the four generation models are inconsistent,  We take eight frames of each video for training and testing and center-crop each frame according to its shortest side, then resize it to 224×224. During the training process, we first resized the image to 256×256 according to bilinear interpolation and then randomly cropped with the size of 224 × 224. Some data augmentation, such as Jpeg compression, random horizontal flipping, and Gaussian blur, are also used. 

\textbf{Model hyperparameters.} Our model consists of two layers of transformer and an MLP
head is an adaptation of the sequence to sequence trans-
former, where the dimension and feature dimension of MLP is 768, path\_size is 32, the dropout parameter is 0.1, and the number of attention heads is 4.

\textbf{Baseline.} (\romannumeral1) CNNDet~\cite{wang2020cnn} detects images generated by GAN. Through a large amount of data enhancement, the model has generalizability on  images generated by various GANs. (\romannumeral2) DIF~\cite{sinitsa2023deep} is a few-shot learning method for image synthesis. (\romannumeral3) DIRE~\cite{wang2023dire} proposes a new image representation method called Diffusion Reconstruction Error (DIRE), which measures the error between the input image and its reconstruction via a pre-trained diffusion model to detect images generated by diffusion models. (\romannumeral4 ) Lgard~\cite{tan2023learning} uses a pre-trained CNN model as a transformation model to convert images into gradients and used these gradients to present generalized artifacts. (\romannumeral5) I3D~\cite{carreira2017quo} introduces a new Inflated 3D ConvNet that is based on 2D ConvNet inflation. (\romannumeral6) Slow~\cite{feichtenhofer2019slowfast} proposes a dual-path model, but we only adopted the Slow Pathway, which is designed to capture the semantic information provided by a small number of sparse frames. (\romannumeral7) X3D~\cite{feichtenhofer2020x3d} designs a smaller and lighter structure and expanded it in different dimensions to achieve a trade-off between efficiency and accuracy. (\romannumeral8) Mvit~\cite{fan2021multiscale} combines the basic idea of multiscale feature hierarchies with transformer models for video and image recognition. (\romannumeral9) UFDetect(UniversalFakeDetect)~\cite{ojha2023towards} detects generated images through pretrained CLIP:ViT visual encoders and nearest neighbors. (\romannumeral10) EVR~\cite{lin2022frozen} uses the CLIP backbone with frozen parameters and multi layer spatiotemporal Transformer decoder for video action recognition.
\subsection{Robustness}
\label{K}
\begin{figure*}[h]
\centerline{\includegraphics[width=\linewidth]{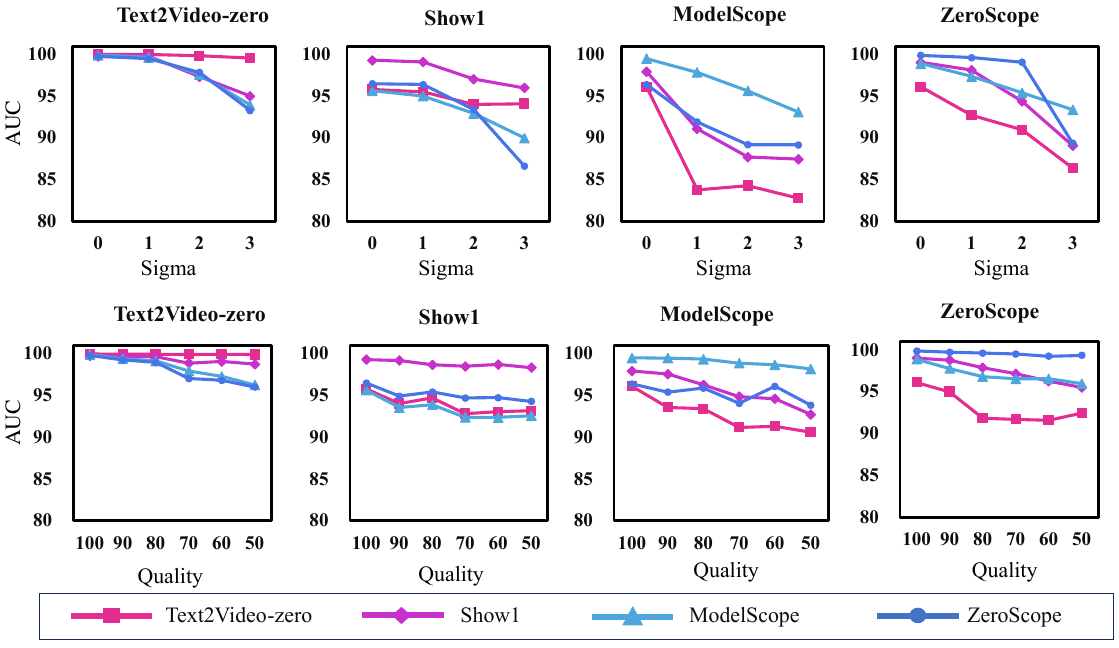}}
\caption{Robustness to unseen perturbations. The upper row represents the robustness of Gaussian blur, and the lower row represents the robustness of JPEG compression. We test the detector trained on different sub-datasets (represented by different colors) on different sub-datasets.}
\label{fig:robu}
\end{figure*}

In practical detection scenarios, the robustness of the detector to unseen perturbations is also crucial. Here, we mainly focus on the impact of two perturbations on the detector: Gaussian blur and JPEG compression. The perturbations are added under three levels for Gaussian blur ($\sigma$ = 1, 2, 3) and five levels for JPEG compression (quality = 90, 80, 70, 60, 50). Figure~\ref{fig:robu} shows the performance of our model trained on different sub-datasets in the face of the above two perturbations. Although experiencing some performance inconsistencies due to differences between training datasets, our method remains effective under all degrees of arbitrary perturbation, with the AUC  higher than 80\%. \\Under any degree of JPEG compression, the AUC of the model trained on any sub-dataset is greater than 90\%. 
\subsection{Different Mapping Functions}
\label{C}
In this section, we examine the impact of different mapping functions. 

The mapping function should have two properties. One is that the distance between features in the feature space is inversely proportional to the similarity between images, and the other is that it is insensitive to spatial artifacts. The second requirement is easy to implement as long as the mapping function is not used to train the fake content detection task. For the first requirement, we explored several different mapping functions, as shown in the table~\ref{tab:mapfunction}. 

First, we discussed the two different tasks of ImageNet and CLIP. Compared to ImageNet's image classification task, CLIP's image encoder trained on text image task alignment is a better mapping function. We considered two reasons: one is the size of the training data, and the other is due to the difference in the task itself; We do not want the model to classify images into specific items or biological categories but rather cluster similar images. At the same time, we compared the impact of different model frameworks: Due to issues such as the number of model parameters, Resnet trained on the CLIP task is not suitable as a mapping function.

Overall, the mapping function largely determines the effectiveness of our approach.

\begin{table*}[t]
  \centering
 \caption{A comparison of the impact of different mapping functions.}
    \begin{tabular}{cccccccccccc}
    \toprule
    \multicolumn{1}{c}{\multirow{3}[6]{*}{\makecell[c]{Mapping\\ Function}}} & \multicolumn{1}{c}{\multirow{3}[6]{*}{\makecell[c]{Pretrain\\ Tasks}}} & \multicolumn{8}{c}{Video generation models}              & \multicolumn{2}{c}{\multirow{2}[4]{*}{Total Avg.}} \\
\cmidrule{3-10}          &       & \multicolumn{2}{c}{Text2Video-zero} & \multicolumn{2}{c}{Show1} & \multicolumn{2}{c}{ModelScope} & \multicolumn{2}{c}{ZeroScope} & \multicolumn{2}{c}{} \\
\cmidrule{3-12}          &       & ACC   & AUC    & ACC   & AUC    & ACC   & AUC    & ACC   & AUC    & ACC   & AUC \\
    \midrule
    ViT-L/14  & CLIP  & 99.50  & 100.00  & 71.50  & 95.83  & 78.25  & 96.13  & 76.50  & 96.13  & \textbf{81.44}  &\textbf{97.02}  \\
    Resnet50 & CLIP  & 96.92  & 99.82  & 52.60  & 80.66  & 57.22  & 80.19  & 50.00  & 67.65  & 64.19  & 82.08  \\
    ViT-B/16  & ImageNet  &98.00  & 99.79  & 65.39 & 87.65 &54.70  & 76.39 & 55.57 & 76.1 & 68.41 &84.98 \\
     Resnet50  & ImageNet  &99.5 &99.96	&54.2 & 81.9	&51.43 & 67.97	&49.93 &58.65	&63.76 & 77.12  \\
    \bottomrule
    \end{tabular}%

  \label{tab:mapfunction}%
\end{table*}%

\subsection{The Impact of Video Frame Number}
\label{D}
Here we discuss the impact of the number of video frames used for detection.
\begin{table*}[t]
  \centering
  
  \vskip 0.15in
\caption{A comparison of the impact of different frame numbers.}
    \begin{tabular}{cccccccccccc}
    \toprule
    \multicolumn{1}{c}{\multirow{3}[6]{*}{\makecell[c]{Training \\ dataset}}} & \multicolumn{1}{c}{\multirow{3}[6]{*}{\makecell[c]{Frame\\ Numbers}}} & \multicolumn{8}{c}{Video generation models}              & \multicolumn{2}{c}{\multirow{2}[4]{*}{Total Avg.}} \\
\cmidrule{3-10}          &       & \multicolumn{2}{c}{Text2Video-zero} & \multicolumn{2}{c}{Show1} & \multicolumn{2}{c}{ModelScope} & \multicolumn{2}{c}{ZeroScope} & \multicolumn{2}{c}{} \\
\cmidrule{3-12}          &       & ACC   & AUC    & ACC   & AUC    & ACC   & AUC    & ACC   & AUC    & ACC   & AUC \\
    \midrule
    Text2Video-zero	&8	&99.50 &100.00	&71.50 &95.83	&78.25 & 96.13	&76.50 &96.13	&81.44&97.02 \\
    Show1	&8	&98.50 &99.77	&97.00 &99.33	&90.25 &97.96	&93.75 &99.08	&94.88 &99.04 \\
    ModelScope	&8	&98.50 &99.94	&80.25 &95.68	&97.00 
    &99.53	&93.75 &98.93	&92.38 &98.52 \\
     ZeroScope	&8	&97.00 &99.81	&75.50 &96.52	&75.75 &96.42	&98.50 &99.91	&86.69 &98.17  \\
    \midrule
    Text2Video-zero	&16	&99.50 &99.99	&60.70 &95.13	&65.55 &93.55	&63.27 &95.38	&72.25 &96.01\\
    Show1	&16	&97.86 &99.69	&95.93 &99.04	&93.73  &98.13	&94.30 &98.88	&95.45 &98.93\\
    ModelScope	&16	&100.00 &100.00	&81.25 &95.41	&98.36  &99.55	&95.36 &99.19	&93.74 &98.53\\
    ZeroScope	&16	&97.43 &99.83	&80.75 &96.10	&82.18 &96.26	&96.93 &99.72	&89.32 &97.97\\
    \midrule
    Text2Video-zero	&24	&99.50 &100.00	&57.57 &95.14	&55.7 &94.10	&60.18 &96.32	&68.23 &96.39\\
    Show1	&24	&97.80 &99.98	&94.36 & 99.10	&89.66 &97.49	&93.73 &98.93	&\textbf{93.88} &98.87\\
    ModelScope	&24	&100.00 &100.00	&74.68 &95.57	&97.00 &99.32	&90.73 &99.30	&90.60 &98.54 \\
    ZeroScope	&24	&98.43 &99.96	&83.32 &97.00	&85.02 &96.56	&97.50 &99.78	&91.06 &98.32\\

    \bottomrule
    \end{tabular}%
    
  \label{tab:NUM}%

\end{table*}%

In our work, we selected 8 frames for generated  video detection for two reasons: on the one hand, the current generation model still has this deficiency in generating long videos. Even the videos currently displayed by Sora can only generate about 300-600 frames, so we believe that 8 frames is an acceptable choice. On the other hand, in order to highlight the effectiveness of our method, we chose a challenging number of 8 frames, and a large number of experiments have also proven that our method indeed performs very well.

We conducted experiments on the impact of frames 8, 16, and 24 on detection, as shown in the table~\ref{tab:NUM}.
 We found that the performance of 16 frames is generally better than that of 8 frames, except for the models trained on the Text2Video zero subdataset, which is consistent with our previous evaluation results of several generation models. Text2Video zero has the worst time continuity. In addition, the effect will become worse at 24 frames. One possible reason is that as the sequence increases, the temporal artifacts of the generated video become more unique, which weakens the model's generalizability.

\subsection{The Impact of Video Frame Sampling Interval.}
\label{E}
In this section, we discussed the impact of sampling interval on detection results. 
In previous experiments, we usually took the first 8 frames of the video for detection. 

Unlike previous experiments, this experiment was conducted on the latest generation model Sora, as Sora has made significant progress in generating frames compared to other generation models, making it easier for us to comprehensively discuss the impact of sampling strategies. 
As shown in the table~\ref{tab:SAM}, facing three different sampling strategies: 1) Uniformly sample 8 frames from the first 32 frames; 2) Take 8 frames evenly from all frames; 3) Take 8 frames evenly from the first 300 frames; Our method did not show a significant decrease, which we attribute to two reasons: 1) The prompts we collected were comprehensive enough to cover the speed of video changes well. 2) The duration of the test video is still around 10 seconds, so the impact of sampling interval is not very significant. For longer generated videos, the performance degradation caused by changes in sampling intervals will be more severe.

\begin{table*}[htbp]
  \centering
   \caption{The impact of different sampling strategies}
    \begin{tabular}{cccc}
    \toprule
    \multicolumn{1}{c}{\multirow{3}[4]{*}{\makecell[c]{Training\\dataset}}} & \multicolumn{1}{c}{\multirow{3}[4]{*}{\makecell[c]{Sampling\\ strategy}}} & \multicolumn{2}{c}{\multirow{2}[2]{*}{Sora}} \\
          &       & \multicolumn{2}{c}{} \\
\cmidrule{3-4}          &       & ACC   & AUC \\
    \midrule
    Text2Video-zero	&Take 8 frames evenly from first 32 frames	&96.87 &99.70\\
    Show1	&Take 8 frames evenly from first 32 frames		&94.79 &98.26	\\
    ModelScope	&Take 8 frames evenly from first 32 frames	s	&94.79 & 98.83	\\
    ZeroScope	&Take 8 frames evenly from first 32 frames		&89.58 &97.92	\\
    \midrule
    Text2Video-zero	&Take 8 frames evenly from all frames	&96.87 &99.48	\\
    Show1	&Take 8 frames evenly from all frames	&94.79 &98.61	\\
    ModelScope	&Take 8 frames evenly from all frames	&94.79 &98.96	\\
    ZeroScope	&Take 8 frames evenly from all frames	&88.54 & 98.31	\\
    \midrule
    Text2Video-zero	&Take 8 frames evenly from first 300 frames	&97.91 &99.65	\\
Show1	&Take 8 frames evenly from first 300 frames	&94.79 &98.48	\\
ModelScope	&Take 8 frames evenly from first 300 frames	&93.75 &98.74	\\
ZeroScope	&Take 8 frames evenly from first 300 frames	&90.62 &98.22	\\

    \bottomrule
    \end{tabular}%
 
\label{tab:SAM}%
\end{table*}%

\subsection{More Details on Sora.}
\label{F}
\subsubsection{Test data on Sora}
Due to regional restrictions, we were not eligible to test Sora. But we tested all of the 48 videos officially displayed by Sora. The test dataset consists of 48 official videos released by Sora and 48 real videos from the original test set, the test on Veo are similar, we tested all of the 14 videos officially displayed by Veo.
\subsubsection{The detection performance of human testers on Sora}
 We invited three testers to evaluate the authenticity of 96 videos (48 real and 48 Sora generated) from the previously mentioned Sora test set. The experimental results are shown in the table~\ref{tab:human}.

The results of DeCoF in testing far exceed the level of human judgment. At the same time, we also need to point out that interviews with testers after testing revealed that their judgment on the videos generated by Sora was artificially high, as there were 4-5 flashy videos in the officially released videos that did not match the results that could appear in the real world, such as pirate ships in coffee cups. Meanwhile, according to the testers, their judgment is based on whether there is irregular movement (such as objects appearing out of thin air) or object threading (such as overlapping basketball and basket) and abnormal distortion (such as human fingers or limbs of moving animals) in the generated video.
\begin{table}[h]
  \centering
  \caption{Compared to the test results of human testers}
    \begin{tabular}{ccc}
    \toprule
    \multicolumn{1}{c}{\multirow{3}[4]{*}{Training dataset}} & \multicolumn{2}{c}{\multirow{2}[2]{*}{Sora}} \\
          & \multicolumn{2}{c}{} \\
\cmidrule{2-3}          & \multicolumn{1}{p{4.04em}}{TP+TN} & ACC \\
    \midrule
    Tester 1	&57(48+9)		&59.37\\
    Tester 2	&66(47+19)		&68.75\\
    Tester 3	&60(48+12)		&62.5\\
    Text2Video-zero	&93  &96.87\\
    Show1	&91		&94.79\\
    ModelScope	&91		&94.79\\
    ZeroScope	&86		&89.58\\
    \bottomrule
    \end{tabular}%

  \label{tab:human}%
\end{table}%
\subsection{Spatial artifacts vs Temporal artifacts.}
\label{G}
\subsubsection{Detect generated videos from temporal artifacts}
Although our method attempts to detect generated videos from temporal artifacts, there is a question: is it possible that CLIP: ViT has differences in the features between the generated video and the real video frames when extracting image features, resulting in better detection performance? In other words, is our method really detect generated videos from temporal artifacts?

To prove this, we repeated the previous probe experiment. As shown in the table~\ref{tab:probe}, our model has also performed well in detecting test data without spatial artifacts. On the other hand, for the second set of test data in the previous probe experiment, our model has also successfully detected it. We believe that repeated stillness in multiple frames is also a type of temporal artifact to some extent.

\subsubsection{Spatial artifacts vs Temporal artifacts.}
In this paper, we attempt to detect generated video  from the perspective of temporal artifacts, but this does not mean that it is impossible to detect generated video  from spatial artifacts. In fact, the difficulty of detecting spatial artifacts in generated videos lies in how to extract more general artifacts, as different generation models generate videos with different resolutions, color brightness, contrast, and so on.
\begin{table}[htbp]
  \centering
  \caption{The proof of our method detect generated videos from temporal artifact }
    \begin{tabular}{ccccc}
    \toprule
    \multicolumn{1}{c}{\multirow{3}[6]{*}{Method}} & \multicolumn{4}{c}{Video generation models} \\
\cmidrule{2-5}          & \multicolumn{2}{c}{  Temporal} & \multicolumn{2}{c}{Spatial} \\
\cmidrule{2-5}          & ACC   & AUC    & ACC   & AUC \\
    \midrule
    I3D~\cite{carreira2017quo}   & 51.17  & 51.61  & 86.67  & 98.61  \\
    Slow~\cite{feichtenhofer2019slowfast}  & 50.04  & 50.37  & 93.81  & 99.51  \\
    X3D~\cite{feichtenhofer2020x3d}   & 50.04  & 54.31  & 78.39  & 92.84  \\
    Mvit~\cite{fan2021multiscale}  & 50.48  & 55.29  & 72.16  & 92.59  \\
    \midrule
    Ours &\textbf{98.46} &\textbf{99.80} &98.46 &99.66\\
    \bottomrule
    \end{tabular}%

  \label{tab:probe}%
\end{table}%
\subsection{Impact Statements}
\label{H}
Our research focuses on detecting generated videos. We developed the DeCoF model and shared our dataset to support detecting generated videos. These works are crucial to protect digital content and prevent the spread of disinformation.
Nevertheless, these tools have the potential to be misused, resulting in a rivalry between video generation and detection technology. We want to advocate for the ethical use of technology and foster research in the creation of tools for verifying media authenticity.
We are confident that this will result in protecting the public from the harms of disinformation, improving the clarity and authenticity  of information distribution, and ensuring the protection of personal privacy for individuals.

\subsection{Limitation}
\label{I}
In this paper, we tackle an overlooked challenge,
i.e., generated video detection, by proposing
a universal detector to deal with the trust crisis
caused by various emerging text-to-video (T2V)
models, especially unseen models.
However, there are various methods of video forgery, such as Deepfake and malicious editing, in which our methods may experience significant performance degradation or inapplicability. Taking Deepfake detection as an example, our method does not add additional detection modules to the face. In addition, Deepfake videos are often locally tampered with, and the generated videos are generated as a whole. Our work focuses on detecting generated videos, which has a more diverse range of forgery scenarios and more flexible forgery techniques. 

\subsection{More Visual Examples}
\label{J}
\begin{figure*}[htbp]
\vskip 0.2in
\begin{center}
 \centerline{\includegraphics[width=\textwidth]{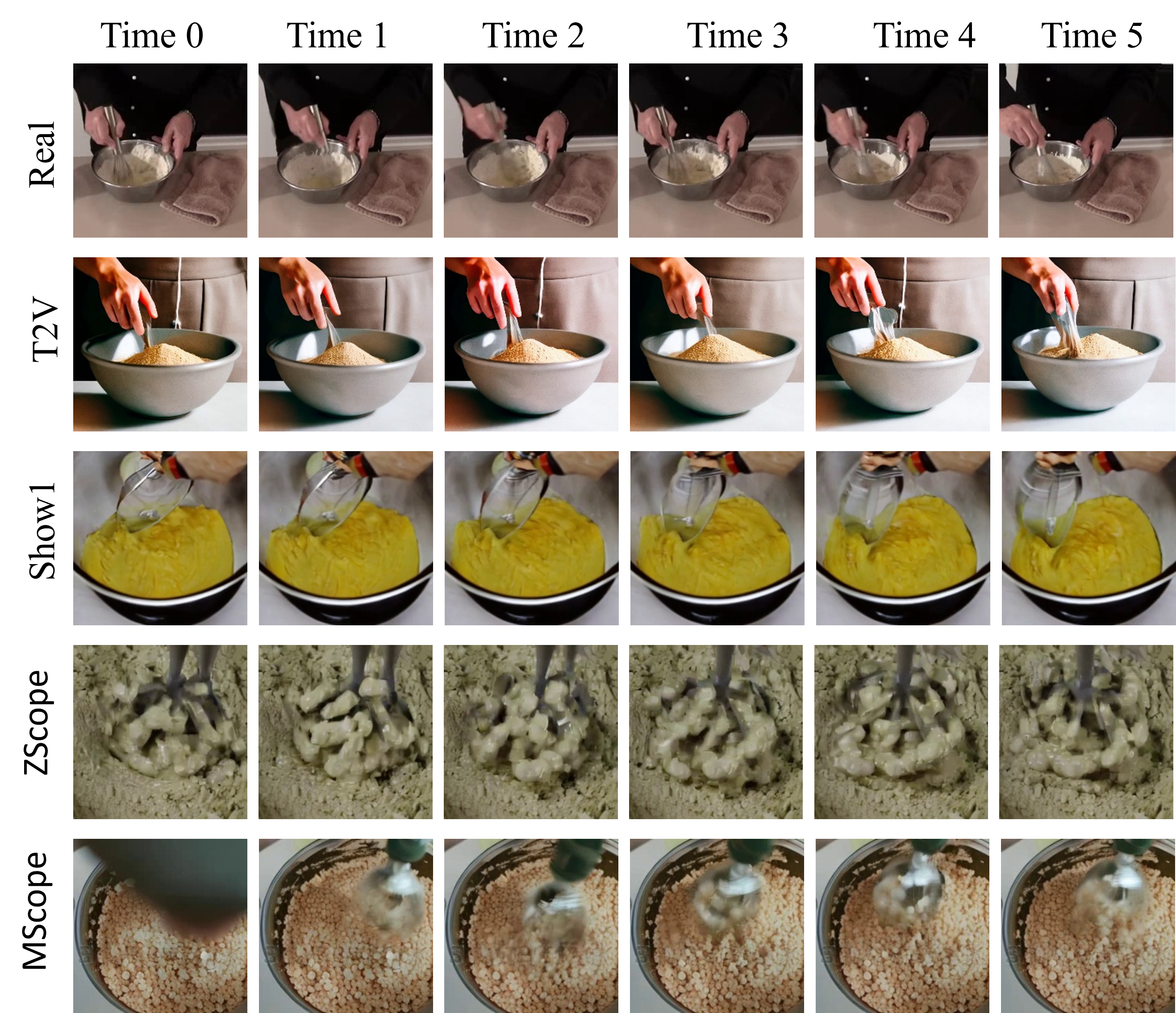}}
 \caption{Real videos and videos generated by four generation models according to prompt: \textbf{A person mixing ingredients in a mixing bowl.}}
\label{EX0}
\end{center}
\vskip -0.2in
\end{figure*}

\begin{figure*}[htbp]
\vskip 0.2in
\begin{center}
\centerline{\includegraphics[width=\textwidth]{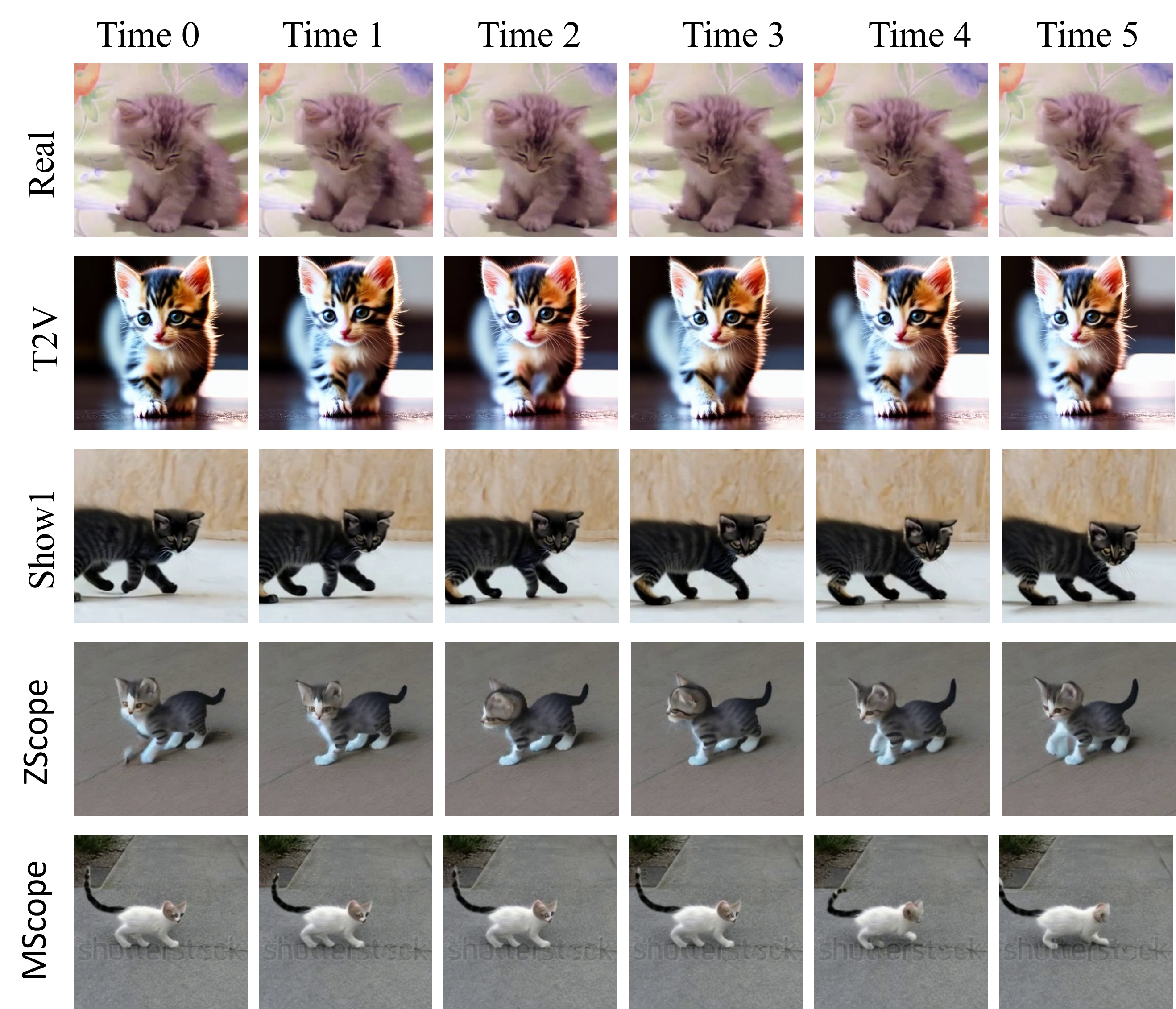}}
\caption{Real videos and videos generated by four generation models according to prompt: \textbf{A kitten moves about.}}
\label{EX1}
\end{center}
\vskip -0.2in
\end{figure*}

\begin{figure*}[htbp]
\vskip 0.2in
\begin{center}
\centerline{\includegraphics[width=\textwidth]{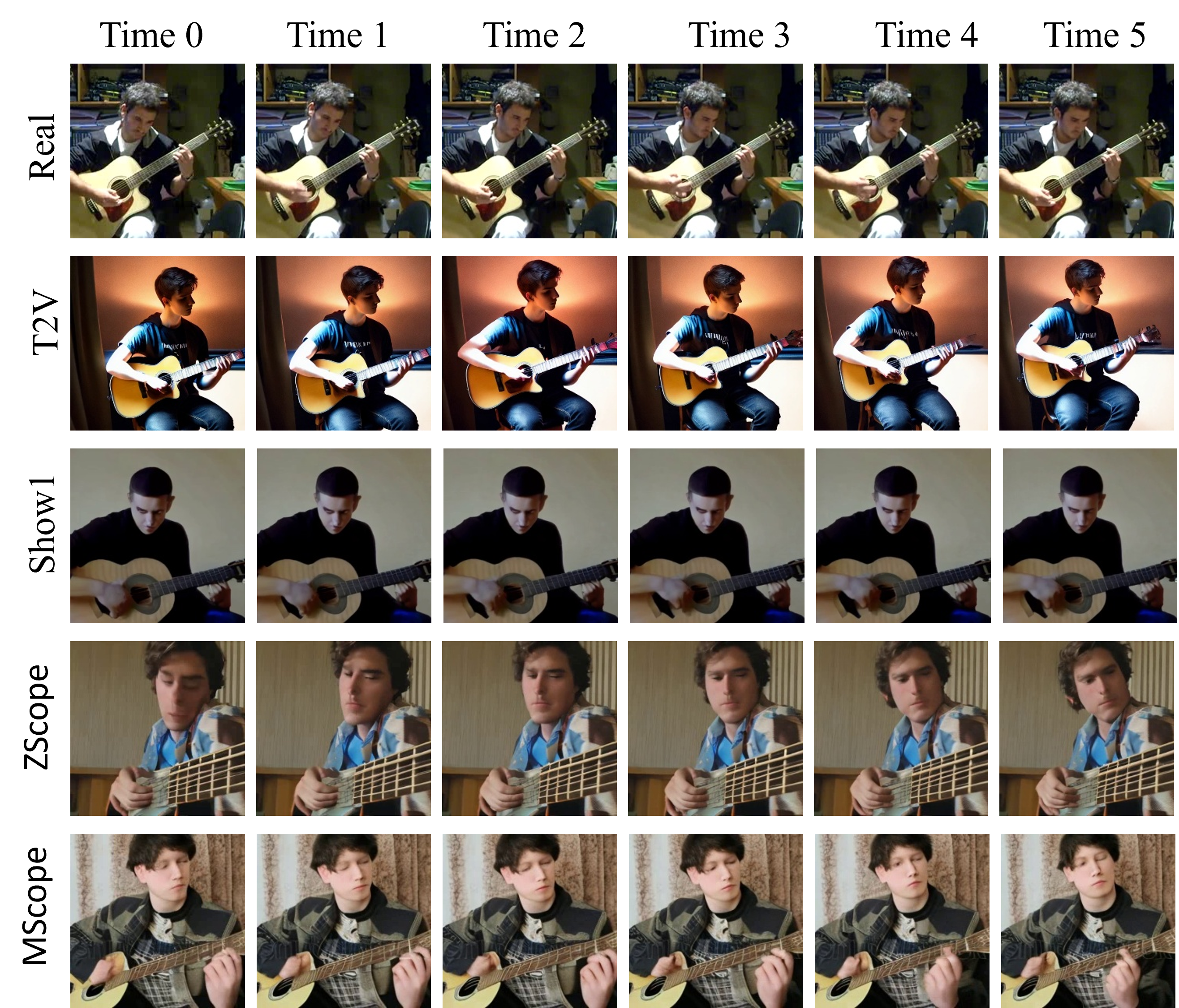}}
\caption{Real videos and videos generated by four generation models according to prompt: \textbf{A young man is seated and playing a guitar.}}
\label{EX2}
\end{center}
\vskip -0.2in
\end{figure*}

\begin{figure*}[htbp]
\vskip 0.2in
\begin{center}
\centerline{\includegraphics[width=\textwidth]{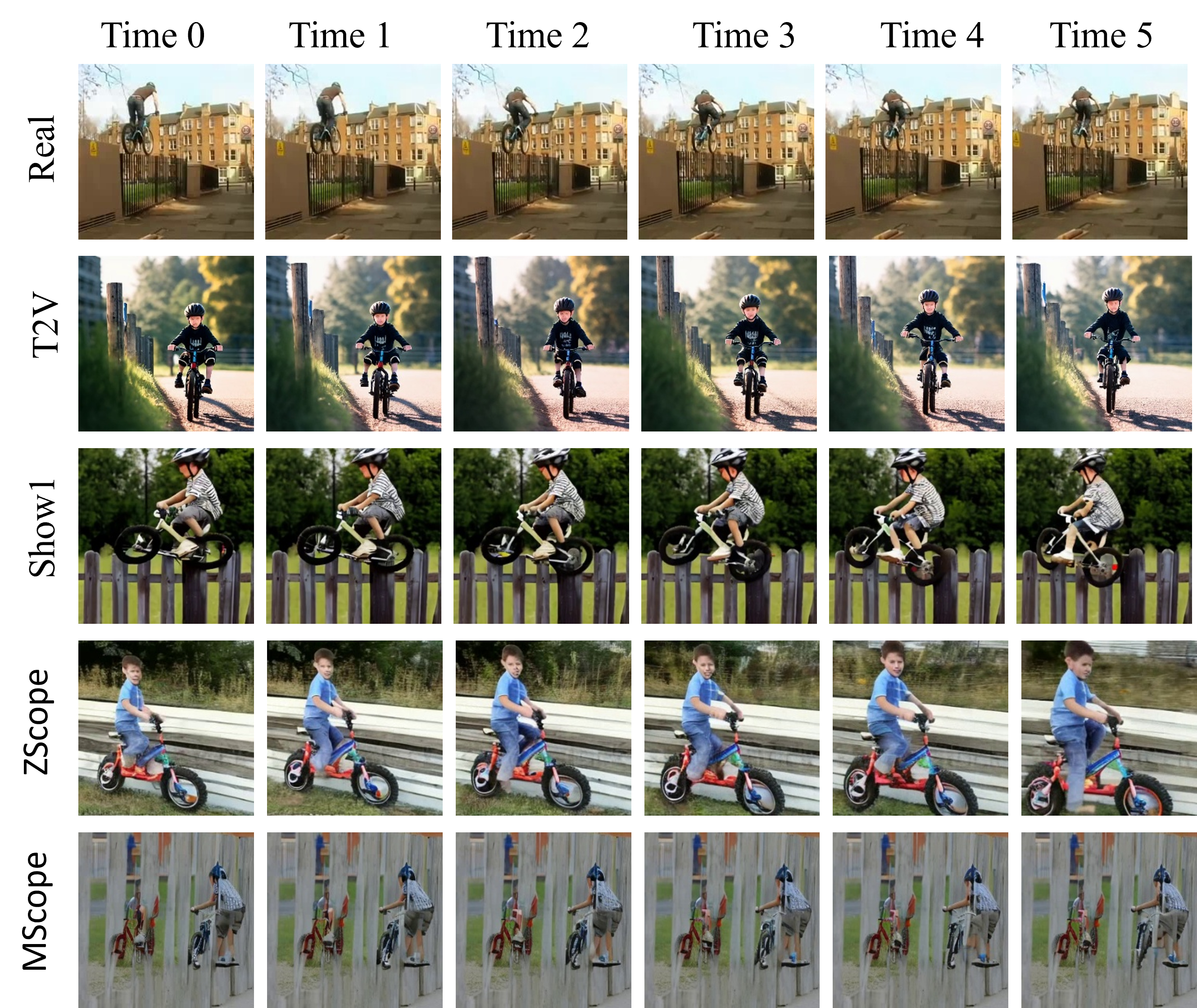}}
\caption{Real videos and videos generated by four generation models according to prompt: \textbf{A boy rides a bike on a fence.}}
\label{EX3}
\end{center}
\vskip -0.2in
\end{figure*}

\begin{figure*}[htbp]
\begin{center}
\centerline{\includegraphics[width=\textwidth]{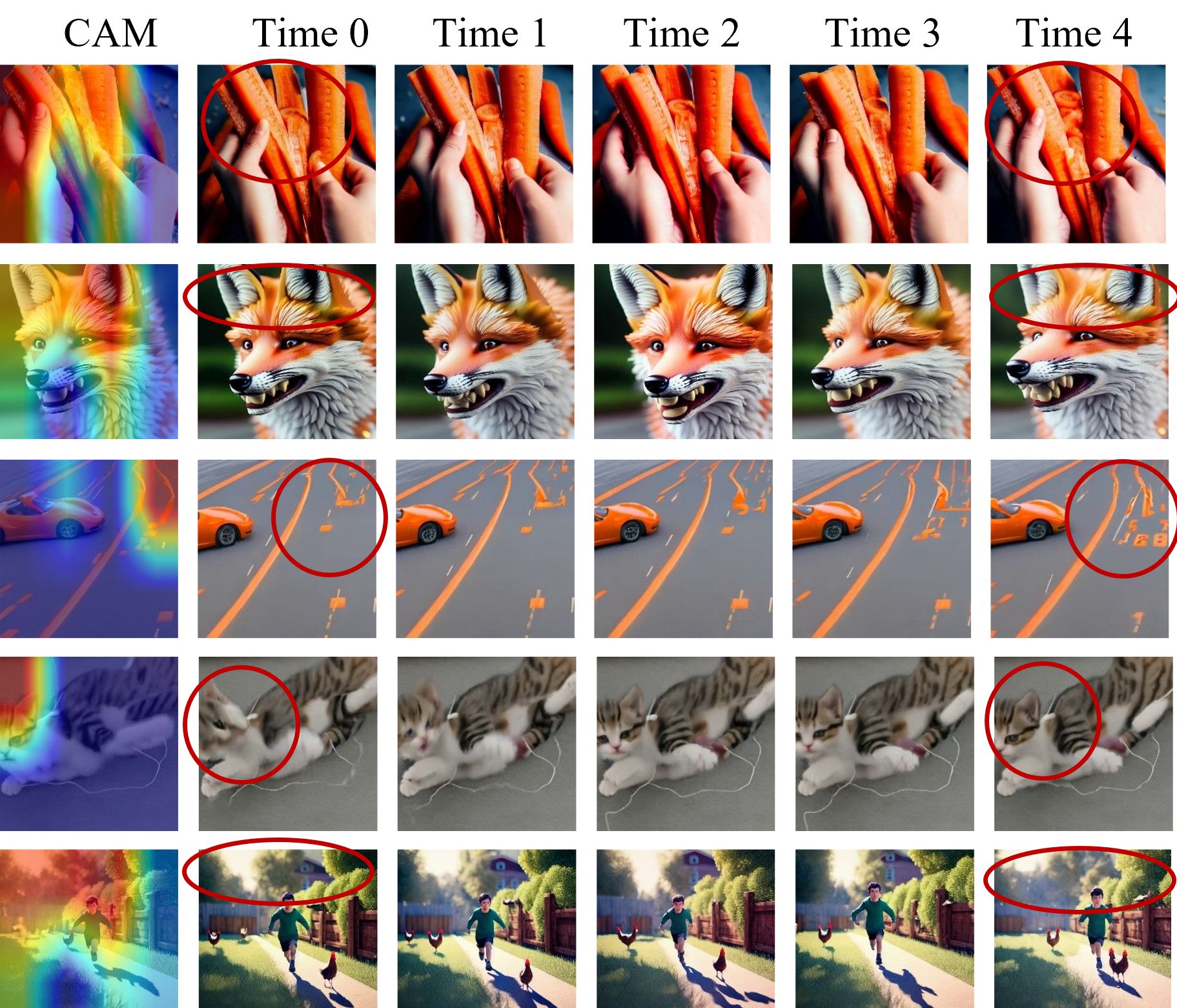}}
\caption{Grad-CAM~\cite{selvaraju2017grad} visualization of our method on different generated videos.}
\label{cam}
\end{center}
\vskip -0.2in
\end{figure*}

Here, we provide more examples of generated videos, as well as Grad-CAM~\cite{selvaraju2017grad} visualization of our method on different generated videos.
Figures~\ref{EX0} to Figures~\ref{EX3} are examples of real videos and generated videos corresponding to prompts. We provide Grad CAM in Figure~\ref{cam} to visualize temporal artifacts in the generated video activated by DeCoF.  It can be concluded from the Figure~\ref{cam} that DeCoF effectively captures the temporal artifacts in the generated video, such as irregular changes in lines on the track, distorted cat heads, abnormal carrot numbers, changes in tree shapes in the background, and so on.

\end{document}